\documentclass[11pt]{article}

\usepackage[preprint]{acl}

\usepackage{times}
\usepackage{latexsym}

\usepackage{hyperref}
\usepackage{algorithm}
\usepackage[noend]{algpseudocode}  %

\usepackage[T1]{fontenc}
\usepackage[utf8]{inputenc}
\usepackage {booktabs}
\usepackage{microtype}

\usepackage{inconsolata}

\usepackage{graphicx}

\usepackage{bm}

\usepackage{amsthm, amssymb}
\usepackage{colortbl}
\usepackage{booktabs}
\usepackage{tabularx}
\usepackage{multirow}
\usepackage{array}

\usepackage[most]{tcolorbox}

\usepackage{amsmath}
\usepackage{amssymb}
\usepackage{enumitem}

\usepackage{graphicx}
\usepackage{subcaption}
\usepackage{dblfloatfix}

\newcommand{\fillin}[1]{\textcolor{black}{#1}}

\title{Where LLM Annotators Fail:\\ Label-Free Learning on Graphs with LLMs}
\author{
Safal Thapaliya, Jiatan Huang,  Chuxu Zhang$^{\dag}$ \\
University of Connecticut, USA \\
\texttt{\{safal.thapaliya,jiatan.huang,chuxu.zhang\}@uconn.edu}\\
{$^{\dag}$Corresponding Author}
}

\newcommand{\methodname}{\textsc{CANE}}     %
\newcommand{\locle}{\textsc{LoCLE}}
\newcommand{\llmgnn}{\textsc{LLM-GNN}}
\newcommand{\dma}{\textsc{DMA}}

\definecolor{promptheader}{HTML}{2F5EA8}
\definecolor{promptbg}{HTML}{EAF1FA}
\newtcolorbox{promptbox}[1][]{
  enhanced, breakable,
  colback=promptbg, colframe=promptheader,
  colbacktitle=promptheader, coltitle=white,
  fonttitle=\bfseries\sffamily,
  title={#1},
  boxrule=0.8pt, arc=3pt,
  left=7pt, right=7pt, top=6pt, bottom=6pt,
  toptitle=3pt, bottomtitle=3pt,
}
\newcommand{\ansslot}[1]{\fbox{\,#1\,}}

\definecolor{ourrow}{HTML}{E8EEF7}      %
\definecolor{grouphdr}{HTML}{EFEFEF}    %
\newcommand{\val}[2]{$#1_{\,\scriptscriptstyle\pm #2}$}            %
\newcommand{\valb}[2]{$\mathbf{#1}_{\,\scriptscriptstyle\pm #2}$}  %
\newcommand{\valu}[2]{$\underline{#1}_{\,\scriptscriptstyle\pm #2}$} %
\newcommand{\na}{\textemdash}                                     %

\usepackage{xcolor}
\usepackage{tikz}
\usepackage{xspace}

\definecolor{caneOrange}{HTML}{F26A21}
\definecolor{caneBlue}{HTML}{254FB8}
\definecolor{canePurple}{HTML}{7A35B4}
\definecolor{caneGreen}{HTML}{2E6B2E}
\definecolor{caneTeal}{HTML}{1B6B73}

\DeclareRobustCommand{\stagecircle}[2]{%
  \tikz[baseline=(char.base)]{%
    \node[
      circle,
      fill=#1,
      text=white,
      inner sep=0pt,
      minimum size=1.35em,
      font=\bfseries\footnotesize
    ] (char) {#2};%
  }%
}

\newcommand{\stageone}{\stagecircle{caneBlue}{1}\xspace}
\newcommand{\stagetwo}{\stagecircle{caneOrange}{2}\xspace}
\newcommand{\stagethree}{\stagecircle{canePurple}{3}\xspace}
\newcommand{\stagefour}{\stagecircle{caneGreen}{4}\xspace}
\newcommand{\stagefive}{\stagecircle{caneTeal}{5}\xspace}

\begin{document}
\maketitle

\begin{abstract}
Node classification on graphs often requires labeled nodes, yet obtaining labels at graph scale is expensive.
When node attributes contain semantic content, such as paper abstracts, web pages, or product descriptions, large language models (LLMs) can provide low-cost supervision by annotating a small subset of nodes.
However, these LLM-generated labels are noisy, and existing label-free graph learning methods usually treat this noise as either global or class-conditional.
We find that LLM annotation errors are not only class-dependent but also region-dependent: within the same class, reliability can vary sharply across feature-space clusters.
In light of this, we propose \textbf{C}luster-\textbf{A}ware \textbf{N}oise \textbf{E}stimation (\textbf{\methodname{}}), a label-free learning framework that estimates cluster-conditional LLM reliability without ground truth labels, and uses this estimate to decide which pseudo-labels to trust, and which labels to correct.
Across various graph benchmarks and GNN backbones, \methodname{} improves over the strongest label-free baselines, with the largest gains on datasets exhibiting stronger cluster-conditional noise. The code is available at \url{https://github.com/thapaliya19/CANE}.

\end{abstract}

\section{Introduction}
\begin{figure*}[t]
    \centering
    \includegraphics[width=\textwidth]{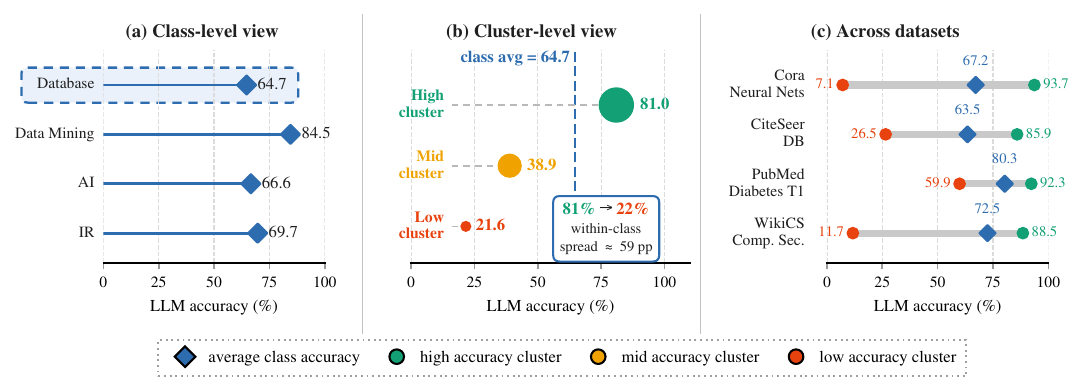}
    \caption{
    \textbf{A single class-level accuracy hides large gaps between clusters of the same class.}
    (a) The class-level view gives each DBLP class one average LLM accuracy.
    (b) Splitting one class into feature-space clusters reveals wide within-class variation in LLM accuracy.
    (c) Low-to-high cluster reliability gaps are consistently observed across different graph benchmarks.
    }
    \label{fig:motivation}
\vspace{-12pt}
\end{figure*}

Graphs are a common representation for relational data in which each node is associated with attributes like descriptive content, metadata, or other features.
In many graphs, these attributes contain semantic information. For example, papers have titles and abstracts, products have descriptions, web pages have text, and users or entities may be associated with profiles or records \cite{yan2023cstag, jin2024llmgraph}.
Such attributes make large language models (LLMs) useful beyond direct text classification: an LLM can read the node content and provide a candidate label when human annotation is unavailable or expensive \cite{he2024tape, wang2025tans}.
This creates a label-free route to graph learning, especially when manual labeling does not scale to large graphs \cite{chen2024llmgnn}.

Recent label-free classification methods build on this idea by using LLMs as annotators rather than as end-to-end predictors \cite{chen2024llmgnn, zhang2025locle, sheng2025dma}.
Instead of querying an LLM for every node, they select a small subset of nodes, ask the LLM to assign labels from the task label space, and train a graph neural network (GNN) on the resulting supervision.
This design combines the semantic knowledge of LLMs with the relational inductive bias of GNNs: the LLM provides inexpensive initial supervision, while the GNN propagates information through graph structure to classify remaining nodes \cite{kipf2017gcn, velickovic2018gat}.

The effectiveness of the above pipeline depends on a question that is often hidden inside the annotation step: \textit{when should an LLM-generated label be trusted?}
LLM annotations are inexpensive, but they are not uniformly reliable.
Once an incorrect label enters a graph learning pipeline, its effect can spread through message passing, pseudo-label expansion, and iterative correction \cite{zhang2025locle}.
Existing label-free methods reduce this risk using confidence filtering \cite{chen2024llmgnn}, class-level correction \cite{sheng2025dma}, or refinement heuristics \cite{zhang2025locle}, but they typically represent LLM reliability at a coarse granularity: either as a global property of the annotator or as a class-conditional confusion pattern.

Our central observation is that this perspective is overly coarse-grained.
In graphs with semantic node attributes, LLM reliability can vary substantially not only across classes, but also across feature-space regions within the same class.
Figure \ref{fig:motivation} illustrates this pattern on DBLP \cite{tang2008dblp}: the "Database" class has an average LLM annotation accuracy of 64.7\%, but its cluster-level accuracy ranges from 81.0\% in one region to 21.6\% in another.
A class-level estimate therefore assigns the same reliability to regions where the LLM behaves very differently, over trusting unreliable clusters and under-using reliable ones.
Appendix \ref{app:diag} shows that this pattern appears across multiple benchmarks and is not an artifact of the clustering procedure.

This observation suggests that label-free graph learning should estimate how LLM errors vary across local regions of the graph, rather than only how often one class is confused with another on average.
The challenge is that such a cluster-conditional estimate appears to require ground-truth labels: for each cluster and class, one would need to know how often the LLM predicts each label given the true class.
We address this challenge with \textbf{C}luster-\textbf{A}ware \textbf{N}oise \textbf{E}stimation (\textbf{\methodname{}}), a label-free framework that estimates local LLM reliability from self-supervised graph representations.
\methodname{} leverages these reliability estimates throughout the learning pipeline to filter pseudo-label acceptance, and adaptively determine which labels to correct during iterative refinement.
Our contributions are as follows:
\begin{itemize}
    \item  We identify cluster-conditional LLM annotation noise as a failure mode in label-free learning on graphs with semantic node attributes. Even within the same true class, LLM reliability can differ substantially across feature-space regions.
    \item We propose \methodname{}, a label-free node classification framework that estimates a per-cluster transition matrix from self-supervised graph embeddings and LLM annotation statistics, without using ground-truth labels.
    \item Across various graph benchmarks and GNN backbones, \methodname{} improves over the strongest label-free baseline, with the largest gains on datasets that exhibit stronger cluster-conditional noise.
\end{itemize}

\section{Methodology}
\label{sec:method}

\begin{figure*}[t]
\centering
\includegraphics[width=\textwidth]{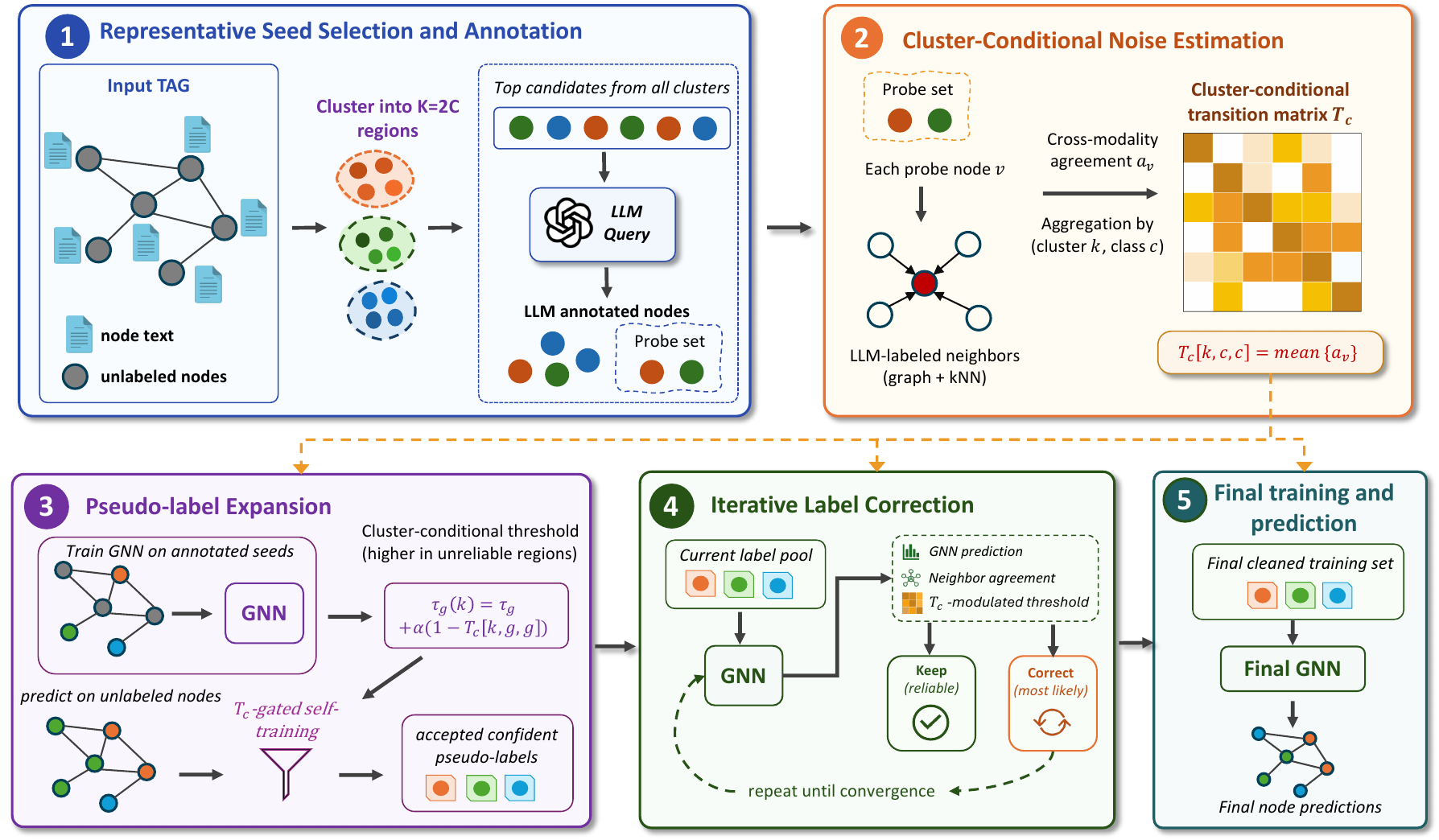}
\caption{Illustration of \methodname{} pipeline. (1) A representative seed set is selected, which is used to (2) estimate a transition matrix $T_c$. The estimated $T_c$ then guides (3) \textit{pseudo-label expansion} and (4) \textit{cluster-conditional iterative label correction}, after which (5) a final GNN is trained on the refined labels to produce node predictions.}
\label{fig:pipeline}
\vspace{-12pt}
\end{figure*}

\subsection{Problem Setup}
\label{sec:problem-setup}
In this work, we study \emph{label-free} node classification on text-attributed graphs (TAG), where no ground-truth labels are available during training.
Given a graph $G=(V,E)$,  where \(V\) and \(E\) denote the node and edge sets, each node $v \in V$ is associated with a semantic text description $t_v$ and an unobserved true label $y_v \in \{1,\ldots,C\}$, where \(C\) is the number of classes in \(G\).
The goal is to assign a class to every node in $V$ without access to any ground-truth label.
Given a query budget $B$, a method selects a subset $S\subseteq V$ with $|S|\le B$ and queries the LLM once per node, obtaining a noisy annotation
$\hat y_v\in\{1,\dots,C\}$
for each $v\in S$.
A GNN is then trained on these annotated seeds and used to classify the remaining nodes.
Two properties make the setting hard.
The supervision is \emph{scarce}--the budget covers only a small fraction of the graph ($B\ll|V|$)--so labels must be propagated well beyond $S$.
And it is \emph{noisy}: the LLM's error rate is not uniform across classes but varies sharply between regions of the graph (Figure~\ref{fig:motivation}).
\methodname{} is organized around both challenges: extending scarce labels across the graph while accounting for where the annotator errs, as we describe next.

\subsection{Overview}
The difficulty identified above, that the LLM errs at rates which vary across
graphs, points to its own remedy: if we knew \emph{where} the annotator is
reliable, we could trust its labels more in those regions and less elsewhere.
\methodname{} makes this knowledge concrete. From the seed annotations alone it
estimates a cluster-conditional measure of LLM reliability, $T_c$, computes it
once, and lets that single estimate govern how the noisy labels are spread across
the graph and how they are revised. Figure~\ref{fig:pipeline} illustrates the five
stages of this method.

The proposed \methodname{} begins by choosing which nodes to annotate.
As no labels yet exist to indicate where the LLM can be trusted, it selects a budget of representative seeds by clustering the graph in feature space (\stageone) and queries the LLM on them.
A leading fraction of those seeds is annotated first as a \emph{probe}, from which \methodname{} estimates $T_c$ (\stagetwo): the one quantity carried into the stages that follow.
Because the probe is drawn from seeds the method already pays for, and selection never consults reliability, this estimate is obtained within the query budget and without any ground-truth label.

The estimate then steers the two stages that assemble the final label set.
During  \emph{pseudo-label expansion} (\stagethree), $T_c$ raises the bar for accepting a model prediction in regions where the annotator is unreliable, so supervision spreads outward from the seeds without absorbing local noise.
During \emph{iterative label correction} (\stagefour), $T_c$ makes the pipeline more reluctant to overwrite a label that already sits in a locally trustworthy region.
A graph neural network trained on the resulting label set produces the final node predictions (\stagefive).
The rest of this section details each stage of the \methodname{} pipeline.

\subsection{Representative Seed Selection and Annotation}
\label{sec:method-annotate}
Under a fixed query budget, the first question is which nodes should be annotated by the LLM.
The choice is important because the seed set provides the only direct supervision from the annotator. It initializes the GNN, drives pseudo-label expansion, and anchors the subsequent correction loop. At the same time, it must be selected blindly from graph structure alone, before any labels are available to indicate where the annotator is reliable. A good seed therefore needs two properties at once: it should represent a meaningful region of the graph while being likely to receive a correct LLM label.

Prior work shows these properties coincide: LLM annotation accuracy is highest for nodes near the centers of feature-space clusters, whose content is most typical of their region~\citep{chen2024llmgnn}.
\methodname{} exploits this through subspace clustering~\citep{zhang2025locle}.
Nodes are embedded with a self-supervised GraphMAE2 encoder~\citep{hou2023graphmae2} and grouped into $K=2C$ feature-space regions, writing $k_v$ for the cluster of node $v$. Within each region, the densest, most central nodes are taken as seeds and annotated by the LLM (Figure~\ref{fig:pipeline}, step~\stageone).
The same partition carries genuine class structure ($63$--$80\%$ single-class purity; see Appendix~\ref{app:clusterability} for details), so each cluster doubles as a unit for the per-cluster reliability estimate that follows.
We allocate the full budget  $B=50C$ to these seeds.
To estimate that reliability without spending extra queries, \methodname{} annotates the first $\rho=0.4$ fraction of the seeds ahead of the rest and designates them a \emph{probe set} $\mathcal{P}\subset\mathcal{S}$. The matrix $T_c$ is estimated from $\mathcal{P}$ alone (\S\ref{sec:method-estimate}) before the remaining $(1-\rho)B$ seeds are annotated to complete $\mathcal{S}$.
The probe reuses labels already collected by the method, so it incurs no additional cost.

\subsection{Cluster-conditional Noise Estimation}
\label{sec:method-estimate}

Once the probe nodes are annotated, the next stage measures how reliable those labels are in each region of the graph: the quantity that will later govern both expansion and correction. \methodname{} records this reliability as a cluster-conditional transition matrix $T_c$ over the same $K=2C$ clusters used for selection (Figure~\ref{fig:pipeline}, step~\stagetwo):
\(
T_c[k,i,j]=P(\hat y=j\mid y=i,\ \mathrm{cluster}=k)\in[0,1]^{K\times C\times C}.
\)
The diagonal entry  $T_c[k,c,c]$ is the probability that the LLM correctly labels a class-$c$ node in cluster $k$, and whose cluster-average $T=\tfrac1K\sum_k T_c[k]$ is the coarser, class-conditional matrix modeled in prior work~\citep{sheng2025dma}. The challenge is that this definition conditions on the true label $y$, which is unavailable in the label-free setting. We therefore estimate the diagonal entries using only LLM annotations on the probe set $\mathcal{P}$. During downstream use, where the true class is also unknown, the diagonal value corresponding to the node’s assigned class is treated as the reliability score for that assignment.

A natural first attempt is to gauge reliability from how \emph{concentrated} the LLM labels are within a cluster, reading a sharply peaked label distribution as a confident and reliable annotator. But concentration conflates two unrelated situations. A cluster that genuinely spans several true classes draws scattered labels even from a perfect annotator, while a single-class cluster labeled by a noisy one scatters them too. As a result, concentration cannot distinguish annotation noise from genuine class mixing, and on the partially mixed clusters common in real graphs (Appendix~\ref{app:clusterability}), it tends to underestimate reliability.

\methodname{} instead asks whether each LLM label is \emph{corroborated} by independent views of the same node. For a probe node $v\in\mathcal{P}$, let $\mathcal{N}(v)$ gather its annotated graph neighbors together with its nearest neighbors in feature space, and define the cross-modality agreement:
\begin{equation}
a_v = \frac{1}{|\mathcal{N}(v)|}\sum_{u\in\mathcal{N}(v)} \mathbf{1}[\hat y_u = \hat y_v],
\label{eq:tc-agreement}
\end{equation}
as the fraction of those neighbors the LLM labeled the same way as $v$. A label that two independent views: graph structure and feature similarity, agree on is more likely correct. We therefore average $a_v$ over the probe nodes of each (cluster, class) pair to obtain the label-free reliability estimate:
\[
R[k,c]=\operatorname{mean}\{a_v : v\in\mathcal{P},\, k_v=k,\, \hat y_v=c\},
\]
which is assigned to the diagonal entry $T_c[k,c,c]=R[k,c]$, with the remaining probability mass distributed uniformly across the off-diagonal entries.
Only the diagonal is estimated and used downstream; the uniform
off-diagonal mass serves only to make each row a distribution, so in
practice $T_c$ acts as a $K\!\times\!C$ table of cluster-conditional
reliabilities.
When a (cluster, class) pair contains too few probe nodes, the estimate backs off to the cluster-level and then global mean agreement.
Because agreement tracks whether a label is \emph{right} rather than whether a region is single-class, the estimate stays calibrated precisely where concentration fails. For example, on \textsc{PubMed}, whose clusters are only $50$--$77\%$ single-class, a concentration-based estimate under-reads reliability by ${\sim}37$pp, whereas the agreement estimate recovers the true ${\sim}88\%$ LLM accuracy. The estimate draws on the probe annotations and no ground-truth labels, yet captures how the annotator's errors shift from region to region rather than averaging them into a single class-level matrix.

\subsection{Pseudo-label Expansion}
\label{sec:method-refine}

Pseudo-label expansion is the first stage guided by $T_c$, addressing the limited supervision available from the $B=50C$ annotated seeds. Since these seeds cover only a small portion of the graph, a GNN trained on them alone cannot adequately capture the full node distribution. A common solution is self-training: iteratively enlarging the training set with the model’s own high-confidence predictions. However, this process is inherently risky. Incorrect pseudo-labels can propagate through later rounds of training and expansion, reinforcing their own errors. This problem is especially severe under a single class-level confidence threshold, which applies the same acceptance criterion everywhere despite the fact that a class may be reliable in some graph regions and unreliable in others.

\methodname{} first trains a GNN on the LLM-labeled seeds under early-learning regularization~\citep{liu2020elr}, which discourages the network from memorizing noisy labels in later training stages.
Letting $\hat p_v\in\Delta^{C-1}$ denote the predicted class distribution for node $v$, the training loss is $\mathcal{L}=\mathcal{L}_{\mathrm{ELR}}(\hat p_v,\hat y_v)$.
An alternative would be to incorporate the noise estimate directly into the loss through forward correction~\citep{patrini2017forward}, as done in \dma{}~\citep{sheng2025dma}.
\methodname{} deliberately avoids this because forward correction relies on a single transition matrix and therefore only uses the cluster-averaged $T$, discarding the local structure encoded in $T_c$.
In practice, this not only fails to improve performance but reduces mean accuracy by $1.9$pp across the five benchmarks (GCN, 5 seeds), likely because a global matrix under-corrects the most difficult clusters (Appendix~\ref{app:diag}).

\methodname{} instead applies $T_c$ where it can stay cluster-conditional, at the pseudo-label admission threshold (Figure~\ref{fig:pipeline}, step~\stagethree).
For an unlabeled node $v$ in cluster $k_v$ with GNN-predicted class $g_v$, it replaces the fixed class-level threshold $\tau_{g_v}$ with a region-adjusted one:
\begin{equation}
\tau_{g_v}(k_v) = \tau_{g_v} + \alpha\bigl(1 - T_c[k_v,g_v,g_v]\bigr),
\label{eq:tc-gate}
\end{equation}
and admits the pseudo-label only when the prediction clears it.
Where the LLM is locally reliable for class $g_v$, the estimated $T_c[k_v,g_v,g_v]$ is high and the threshold stays near its base value, so the region expands freely.
Where reliability is low, the threshold tightens and admits only the most confident predictions.
\fillin{
We emphasize that the gate uses only a single diagonal entry of $T_c$, $T_c[k_v,g_v,g_v]$, which estimates the reliability of the predicted class within the node's cluster, and discards the rest of the transition structure. Other measures of local annotation quality could play the same role, as demonstrated by the ablation in Section~\ref{sec:ablations}. The key contribution of $T_c$ is therefore to make the pseudo-label admission threshold \emph{cluster-aware}.
}
Expansion is therefore governed not by model confidence alone but by the local trustworthiness of the supervision that produced it.

\subsection{Iterative Label Correction}
\label{sec:method-ilc}
Expansion enlarges the label pool, but neither the labels it adds nor the seed labels it retains are guaranteed correct.
A label may be inconsistent with what the trained GNN now predicts, or unsupported by the labels around it.
If left uncorrected, these errors can accumulate over successive rounds of retraining and expansion. The final stage therefore revisits the label pool, iteratively retraining the GNN and updating labels whose supporting evidence has changed until the pool stabilizes.

The key question is when a disagreement between the GNN and a node's current label should trigger a correction.
A uniform rule treats all disagreements equally, even though their meaning varies across regions of the graph.
Where the LLM is reliable, the current label is probably right and the GNN is the one to doubt; where it is noisy, both the label and the GNN trained on it may be wrong, so the disagreement is weaker evidence for a change.
\methodname{} accordingly conditions each revision on the local reliability $T_c$ (Figure~\ref{fig:pipeline}, step~\stagefour).

Each round retrains the GNN on the current pool and compares its prediction at every labeled node with that node's label.
If the prediction agrees with the current label, the label is kept unchanged. Otherwise, \methodname{} updates the label to the GNN prediction only when sufficient support exists among the node’s labeled neighbors. Specifically, the neighbors agreeing with the GNN prediction must exceed the cluster-dependent threshold: \(
\theta_v = \theta_0 + \beta\,T_c[k_v,\bar y_v,\bar y_v],
\)
where $k_v$ is the cluster of node $v$ and $\bar y_v$  is the current label.
The threshold rises with the reliability of the current label in its region, so a label the LLM annotates dependably is overwritten only under strong neighbor support, while a label in a noisy region yields more readily.
As a result, each labeled node is thus either \emph{kept} or \emph{corrected} in a round.
The loop continues until the labels stabilize, typically requiring only two to four rounds across all benchmarks, with the first round accounting for $83$--$98\%$  of all corrections (Appendix~\ref{app:ilc-conv}).
A final GNN trained on the stabilized label pool then produces the node predictions (Figure~\ref{fig:pipeline}, step~\stagefive).
\fillin{Importantly, a correction does not guarantee improvement: updating a label toward the GNN prediction can move it either toward or away from the true class.
We therefore evaluate whether the correction step is beneficial in aggregate rather than merely measuring how often it changes labels.
Across the ten settings, $56.8\%$ of corrections recover the correct
class, above the corresponding $11$--$50\%$ chance rates, and the
correction step is net-positive with respect to ground-truth accuracy in
nine of the ten settings (Appendix~\ref{app:correction-accuracy}).
}

\section{Experiments}
\label{sec:experiments}

\begin{table*}[!t]
\centering
\small
\setlength{\tabcolsep}{6pt}
\renewcommand{\arraystretch}{1.18}
\begin{tabular}{@{}lccccc@{}}
\toprule

\textbf{Method} & \textsc{CiteSeer} & \textsc{Cora} & \textsc{PubMed} & \textsc{WikiCS} & \textsc{DBLP} \\
\midrule

\rowcolor{grouphdr}\multicolumn{6}{@{}l}{\textit{GAT backbone}}\\
\llmgnn{} (\textsc{FP}) \citep{chen2024llmgnn}  & \val{62.98}{2.62} & \val{69.89}{0.43} & \val{80.63}{0.31} & \val{65.33}{1.63} & \val{68.16}{0.88} \\
\llmgnn{} (\textsc{GP}) \citep{chen2024llmgnn} & \val{67.05}{0.28} & \val{67.72}{1.79} & \val{78.59}{0.46} & \val{63.74}{1.38} & \val{69.79}{1.45} \\
\llmgnn{} (\textsc{RIM}) \citep{chen2024llmgnn}    & \val{65.40}{0.45} & \val{67.41}{0.80} & \val{77.72}{0.87} & \val{64.06}{0.33} & \val{66.27}{1.61} \\
\dma{} \citep{sheng2025dma}      & \val{64.89}{1.36} & \val{69.67}{1.79} & \val{77.31}{2.22} & \val{68.73}{1.62} & \val{65.13}{4.23} \\
\locle{} \citep{zhang2025locle}  & \valu{72.16}{0.91} & \valu{75.23}{1.52} & \valu{81.08}{1.20} & \valu{68.86}{1.44} & \valu{76.36}{0.57} \\
\rowcolor{ourrow}
\textbf{\methodname{}~(ours)}    & \valb{75.11}{0.43} & \valb{75.60}{0.52} & \valb{81.82}{0.80} & \valb{74.91}{1.04} & \valb{77.28}{0.84} \\
\midrule

\rowcolor{grouphdr}\multicolumn{6}{@{}l}{\textit{GCN backbone}}\\
\llmgnn{} (\textsc{FP}) \citep{chen2024llmgnn}  & \val{63.94}{2.92} & \val{71.49}{0.56} & \valu{81.82}{0.85} & \val{67.31}{0.99} & \val{70.14}{0.86} \\
\llmgnn{} (\textsc{GP}) \citep{chen2024llmgnn} & \val{68.91}{0.37} & \val{71.06}{0.61} & \val{80.77}{0.23}  & \val{67.58}{0.65} & \valu{72.95}{0.58} \\
\llmgnn{} (\textsc{RIM}) \citep{chen2024llmgnn}    & \val{65.31}{0.32} & \val{70.36}{0.32} & \val{78.92}{0.23}  & \val{64.79}{0.62} & \val{69.24}{0.16} \\
\dma{} \citep{sheng2025dma}      & \val{63.36}{0.60} & \val{67.08}{1.66} & \val{78.59}{1.59} & \val{66.42}{1.54} & \val{67.65}{2.36} \\
\locle{} \citep{zhang2025locle}  & \valu{73.62}{0.87} & \valb{79.81}{0.57} & \val{81.69}{0.85} & \valb{72.73}{0.82} & \val{72.89}{2.47} \\
\rowcolor{ourrow}
\textbf{\methodname{}~(ours)}    & \valb{75.92}{0.53} & \valu{75.04}{0.48} & \valb{81.85}{0.65} & \valu{72.70}{2.23} & \valb{75.66}{0.86} \\

\bottomrule
\end{tabular}
\caption{Test accuracy ($\%$, 5-seed mean${}_{\pm\text{std}}$) on five
graph benchmarks. Within each backbone block,
\textbf{bold} = best and \underline{underline} = second-best per column.
All methods share the same \texttt{gpt-3.5-turbo} annotator and query budget.}
\label{tab:main}
\vspace{-10pt}
\end{table*}

\subsection{Setup}
\label{sec:exp-setup}

\paragraph{Datasets.}
We evaluate \methodname{} on five label-free TAG benchmarks: three citation networks ( \textsc{CiteSeer}, \textsc{Cora},
\textsc{PubMed}~\citep{sen2008collective}), a Wikipedia web graph (\textsc{WikiCS}~\citep{mernyei2020wikics}), and a publication-venue graph (\textsc{DBLP}~\citep{tang2008dblp}).
These datasets span an order of magnitude in node count and from $3$ to $10$ classes (See Appendix Table~\ref{tab:datastats} for details).

\paragraph{GNN backbones.}
We evaluate two GNN backbones:  GCN~\citep{kipf2017gcn} and
GAT~\citep{velickovic2018gat}. Unless noted otherwise, reported numbers are
means over $5$ random seeds, and ground-truth labels are used only for evaluation.

\paragraph{Baselines.}
We compare against multiple published label-free state-of-the-art methods.
\llmgnn{}~\citep{chen2024llmgnn} pairs active selection with confidence post-filtering. We report its three strongest variants: \textsc{FP}, \textsc{GP}, and \textsc{RIM}, built on FeatProp~\citep{wu2019featprop}, GraphPart~\citep{ma2022partition}, and RIM~\citep{zhang2021rim}.
\dma{}~\citep{sheng2025dma} estimates a class-conditional confusion matrix from synthetic probes and applies forward loss correction~\citep{patrini2017forward}.
\locle{}~\citep{zhang2025locle} iteratively refines labels with graph rewiring but no explicit noise model.
All three run under a matched protocol: the same \texttt{gpt-3.5-turbo} annotator, query budget, and all-node evaluation (See Appendix~\ref{app:setup} for details).

\subsection{Main Results}
\label{sec:exp-main}

Across both backbones, \methodname{} beats \locle{}, the strongest baseline, on
eight of the ten dataset--backbone settings and outperforms the matched-budget
\llmgnn{} and \dma{} on every one (Table~\ref{tab:main}).
\fillin{The comparison with DMA is the most direct: both estimate an LLM confusion matrix and use it for forward correction, but DMA uses a single global matrix rather than a cluster-conditional one. DMA trails \methodname{} on every cell, and Appendix~\ref{app:diag} identifies the mechanism: the global matrix overestimates accuracy in the hardest clusters, causing under-correction where it is most needed. We do not attribute the full gap to cluster conditioning, however, because DMA also lacks the expansion and correction loop shared by \methodname{} and \locle{}. The controlled ablation in Section~\ref{sec:ablations} isolates the noise-model contribution within the same pipeline and finds a smaller effect. Thus, cluster conditioning improves the noise model, but is only one component of \methodname{}'s overall gain.}

The size of the improvement tracks how cluster-conditional a dataset's noise
actually is. It is largest on \textsc{DBLP}, \textsc{WikiCS}, and
\textsc{CiteSeer}, where the LLM's accuracy swings sharply between clusters of the
same class and the label-free estimator stays well-calibrated, so the per-cluster
correction has both the most local structure to exploit and an accurate map of it
(Appendices~\ref{app:diag} and~\ref{app:error-analysis}). \textsc{PubMed} is the
opposite case: its annotations are already accurate and roughly uniform across the
graph, leaving a per-cluster view little to add over a global one, and so
\methodname{} neither helps nor harms. The \textsc{Cora}-GCN shortfall, by
contrast, is structural rather than a failure of noise handling: \locle{} gains
there from Dirichlet-energy graph rewiring: most effective on small, dense,
homophilous citation graphs, which \methodname{} omits, and \textsc{Cora}'s small
graph leaves the estimator little reliable evidence per cluster. The deficit is
narrow and disappears under GAT, where attention plays the role of rewiring, so
combining local reliability with structural rewiring is a natural extension.

\subsection{Annotator Robustness}
\label{sec:exp-annotator}
To test whether cluster-conditional noise is an artifact of weaker
annotators, we swap the \textsc{gpt-3.5-turbo} annotations for
\textsc{gpt-5.4-mini} annotations of the \emph{same} selected nodes and
rerun the pipeline.
Only $T_c$ is re-estimated label-free from the new
annotator's errors. \methodname{} beats \locle{} on eight of the ten
cells, with four margins exceeding \locle{}'s seed standard deviation.
As in the main results, the size of the gain tracks how much local
structure the annotator's errors retain: it is largest on
\textsc{CiteSeer} and \textsc{WikiCS}, where the frontier model still
errs unevenly across clusters of the same class, and narrows on \textsc{DBLP} and \textsc{PubMed}-GAT, whose
annotations are close to clean, while the \textsc{Cora}-GCN deficit of
\S\ref{sec:exp-main} persists.
Absolute accuracy varies with annotator quality, but \methodname{}’s relative gains remain consistent: the method captures where an annotator is reliable or unreliable, rather than adapting to the quirks of any particular annotator.

\begin{table}[h]
\centering
\small
\setlength{\tabcolsep}{6pt}
\renewcommand{\arraystretch}{1.1}
\begin{tabular}{@{}lccc@{}}
\toprule
\textbf{Dataset} & raw LLM & \locle{} & \methodname{} \\
\midrule
\rowcolor{grouphdr}\multicolumn{4}{@{}l}{\textit{GAT backbone}}\\
\textsc{CiteSeer} & 65.38 & \val{64.89}{1.95}  & \valb{72.37}{1.14} \\
\textsc{Cora}     & 66.58 & \val{75.51}{0.21}  & \valb{75.60}{1.01} \\
\textsc{PubMed}   & 86.06 & \val{80.14}{0.27}  & \valb{80.31}{0.05} \\
\textsc{WikiCS}   & 74.27 & \val{68.92}{2.34}  & \valb{75.53}{0.35} \\
\textsc{DBLP}     & 76.93 & \val{77.01}{0.53}  & \valb{77.29}{1.01} \\
\midrule
\rowcolor{grouphdr}\multicolumn{4}{@{}l}{\textit{GCN backbone}}\\
\textsc{CiteSeer} & 65.38 & \val{67.39}{1.04}  & \valb{71.93}{0.30} \\
\textsc{Cora}     & 66.58 & \valb{82.32}{0.65} & \val{76.10}{0.81} \\
\textsc{PubMed}   & 86.06 & \val{79.12}{2.33}  & \valb{81.84}{0.68} \\
\textsc{WikiCS}   & 74.27 & \val{73.52}{1.54}  & \valb{73.92}{1.32} \\
\textsc{DBLP}     & 76.93 & \valb{78.38}{0.05} & \val{77.68}{0.77} \\
\bottomrule
\end{tabular}
\caption{\textbf{Annotator robustness} (\textsc{gpt-5.4-mini}
annotations; test accuracy $\%$,
mean${}_{\pm\text{std}}$). \textbf{Bold} = best per row;
\emph{raw LLM} is the accuracy of the unprocessed annotations.
}
\label{tab:annotator}
\vspace{-15pt}
\end{table}

\subsection{Ablation Study}
\label{sec:ablations}
\begin{figure}[h]
\centering
\includegraphics[width=\columnwidth]{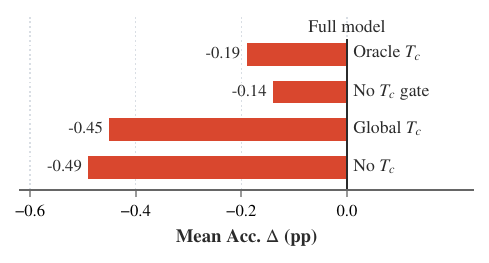}
\vspace{-25pt}
\caption{\textbf{Component ablation}: mean accuracy change (pp) from removing each
method component.}
\label{fig:ablations}
\vspace{-12pt}
\end{figure}

\methodname{}'s use of \(T_c\) rests on three choices: that a noise model
helps at all, that it should be \emph{cluster}-conditional, and that a
\emph{label-free} estimate suffices.
We evaluate each choice through ablations that toggle one component at a time across all ten settings (Figure~\ref{fig:ablations}).
Every
ablation lowers accuracy, and the effects concentrate on the
high-heterogeneity \textsc{gat} cells, exactly where local noise structure
exists; the GCN cells move within seed noise.

\textit{Does \(T_c\) help at all?} Removing it entirely: no expansion gate and
a uniform correction threshold, lowers accuracy, and almost all of that drop
falls on the \textsc{gat} cells. The noise model contributes a meaningful, though modest, portion of the gains.

\textit{Is the benefit specifically cluster-conditional?} Collapsing \(T_c\) into a
single class-conditional matrix removes most of that benefit. The key advantage comes from modeling local noise rather than simply introducing a global noise matrix.

\textit{Does the label-free estimate cost accuracy against an oracle?} A
\(T_c\) built from ground-truth labels performs nearly identically to our estimate, leaving little oracle headroom and supporting the \emph{label-free} claim. Overall, \methodname{} is most effective when strong cluster-conditional structure exists and degrades gracefully when it does not.

\paragraph{Does the gate need a noise model, or only cluster-awareness?}
To test whether a simpler statistic could replace $T_c$ at the admission
gate, we substituted the reliability term $T_c[k_v, g_v, g_v]$ with a
normalized per-cluster variance of the GNN's predictive confidence---a
control that is cluster-aware but carries no noise model. Across all ten
settings the two are indistinguishable: the $0.2$pp difference in mean
accuracy falls well within the $1.3$pp seed standard deviation. This means that at the pseudo-label admission gate, what
matters is that the threshold varies across feature-space regions, not
the particular estimate driving that variation. The structural value of
$T_c$ lies elsewhere in the pipeline: its diagonal tracks ground-truth
cluster purity ($r=0.43$--$0.73$), a $T_c$ built from ground-truth
labels performs nearly identically to the label-free estimate, and
collapsing $T_c$ into a single class-conditional matrix sacrifices
$0.45$ of the $0.49$pp that the noise model contributes overall.

\begin{figure}[h]
\centering
\includegraphics[width=0.85\columnwidth]{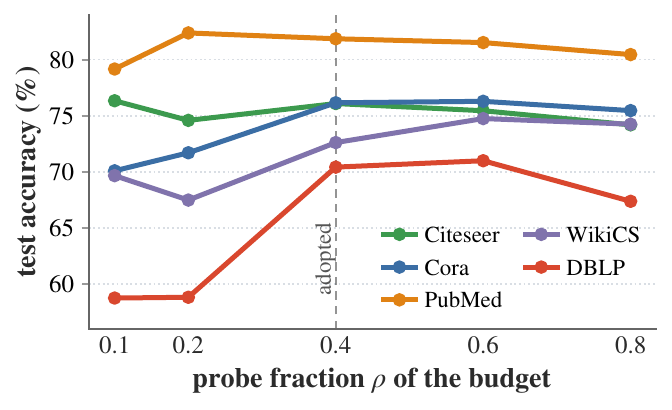}
\vspace{-10pt}
\caption{\textbf{Probe-budget ablation}: test accuracy vs.\ $\rho$,
the share of the budget spent estimating $T_c$ (GCN).}
\label{fig:probe-budget}
\vspace{-12pt}
\end{figure}

\paragraph{Probe budget.}
The probe also serves as the training seed set, so it incurs no additional cost.
The key question is how much probe support the $T_c$ requires to become reliable. Sweeping the
probe fraction $\rho$ (Figure~\ref{fig:probe-budget}), most datasets plateau by
$\rho\approx0.4$ and dip only slightly by $\rho=0.8$, as too few seeds remain for
the rest of the pool. Below $\rho=0.4$, the estimate breaks down on the most challenging dataset, \textsc{DBLP}, where the four-class $T_c$ needs the most support.
We therefore choose $\rho=0.4$, which lies just beyond this transition point across all datasets.

\begin{figure}[h]
\centering
\includegraphics[width=\columnwidth]{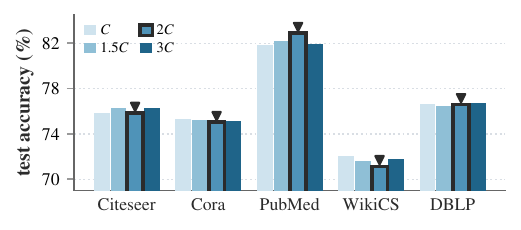}
\vspace{-10pt}
\caption{\textbf{Sensitivity to the number of clusters} $K$ $\in\{C,1.5C,2C,3C\}$.}
\label{fig:ksens}
\end{figure}
\vspace{-12pt}
\paragraph{Number of clusters $K$.}

\methodname{} partitions each graph into $K\!=\!2C$ clusters to condition
$T_c$, trading finer noise resolution against fewer annotations.
Sweeping $K\in\{C,1.5C,2C,3C\}$ (Figure~\ref{fig:ksens}), accuracy barely
moves---the five-dataset mean stays within $0.1$pp.
This is because each cell's
reliability is estimated from neighbor agreement across all nodes, not just the
few probe labels that fall inside it.
So our choice of $2C$ is robust.

\subsection{Budget Sensitivity}
\label{sec:exp-budget}
\begin{figure}[h]
\centering
\vspace{-12pt}
\includegraphics[width=\columnwidth]{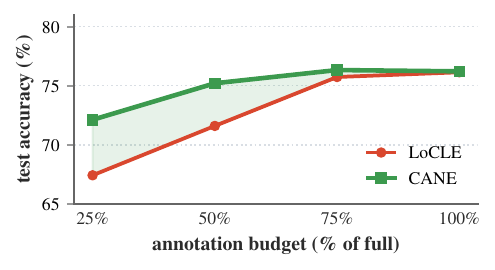}
\caption{\textbf{Budget sensitivity} (GCN, 5-seed mean over the five datasets): test
accuracy of \methodname{} and \locle{} from $25\%$ to $100\%$ of the full
annotation budget.}
\label{fig:budget}
\vspace{-12pt}
\end{figure}

We evaluate performance across annotation budgets ranging from one-quarter to the full per-dataset allocation (GCN, 5 seeds; Figure~\ref{fig:budget}).
The two methods are close at full budget, but the gap widens steadily
as labels grow scarce: \methodname{} degrades gracefully, whereas
\locle{} falls off quickly and, on several datasets, collapses once it
can no longer label enough nodes reliably. \methodname{}'s advantage is therefore largest in the low-budget regime.
This is the regime where reducing annotation cost is most important. As in the main results, \textsc{Cora} remains the exception, where \locle{} is more robust to reduced budgets. Consequently, the overall advantage is primarily driven by the noisier datasets. Per-dataset numbers are in
Appendix~Table~\ref{tab:budget-full}.

\subsection{When to Use CANE?}
\label{sec:deployment}

CANE helps when the budget and the feature-space partition support a
reliable $T_c$, and adds little when they do not. This condition is
measurable in advance, without labels. Before annotating, two quantities
are available: the silhouette of the $K{=}2C$ GraphMAE2 partition, and
the nominal probe density $\rho B/(KC)$, which under our protocol
reduces to $10/C$ and so falls as the label space grows.
\textsc{ogbn-arxiv} fails both ($0.117$; $0.25$), placing it out of
scope. Passing them is necessary but not sufficient: \textsc{Cora}
clears the density check at $1.43$ and is still our weakest setting.
The binding criterion is measured after probing, again label-free---the
\emph{low-support fraction}, the share of (cluster, class) cells that
back off for want of probe evidence (§\ref{sec:method-estimate}). It
orders the outcomes in Table~\ref{tab:deployment}: the three settings
with no low-support cells carry CANE's gains, \textsc{Cora}
($28.6\%$) is the loss, and \textsc{PubMed} ($33.3\%$) is neutral
because its estimate is sparse \emph{and} its $88\%$-accurate
annotations leave a local view little to add. We therefore recommend
deploying CANE below $10\%$ low-support and expecting
class-conditional behaviour above roughly $25\%$. The downside is
bounded by construction: backed-off cells receive stricter admission
thresholds, so the gate never infers cluster-specific trust from
missing evidence, and in the limit CANE reduces to a class-conditional
pipeline rather than failing catastrophically.

\begin{table}[t]
\centering
\vspace{-12pt}
\small
\setlength{\tabcolsep}{5pt}
\begin{tabular}{@{}lccc@{}}
\toprule
Dataset & probes/cell & silh. & low-supp. (\%) \\
\midrule
\textsc{CiteSeer}   & 1.67 & \fillin{0.190} & 0.0  \\
\textsc{Cora}       & 1.43 & \fillin{0.204} & 28.6 \\
\textsc{PubMed}     & 3.33 & \fillin{0.164} & 33.3 \\
\textsc{WikiCS}     & 1.00 & \fillin{0.192} & 0.0  \\
\textsc{DBLP}       & 2.50 & \fillin{0.221} & 0.0  \\
\midrule
\textsc{ogbn-arxiv} & 0.25 & 0.117     & 97.0 \\
\bottomrule
\end{tabular}
\caption{\textbf{Deployment diagnostics.} The low-support fraction,
measured after probing without labels, orders the per-setting outcomes
of Table~\ref{tab:main}; \textsc{ogbn-arxiv} fails both a-priori
screens.}
\label{tab:deployment}
\vspace{-15pt}
\end{table}

\subsection{Cost and Efficiency}
\label{sec:exp-cost}

In label-free graph learning, the dominant cost is the LLM API, not GNN training.
All pipelines use the same annotator and query budget (Appendix~\ref{app:setup}),
so their annotation token cost is identical (Table~\ref{tab:compute}). The budget
sweep of Figure~\ref{fig:budget} therefore doubles as a cost-accuracy curve, on
which \methodname{} reaches any given accuracy at a lower token cost than
\locle{}, especially in low-budget settings where labels are scarce. GNN training is a minor cost by
comparison, and there too \methodname{} is cheaper than \locle{}, whose graph
rewiring and five-stage refinement dominate its runtime.
Beyond GNN training, \methodname{}'s own machinery adds no LLM queries:
$T_c$ is estimated in a single pass over probe seeds the budget already
covers, the GraphMAE2 encoding is a one-time cost shared with
\locle{}'s subspace selection, and expansion and correction converge in
$2.4$--$4.0$ rounds (Appendix~\ref{app:ilc-conv}). The end-to-end
comparison therefore reduces to the annotation cost, which is matched by
construction, and the training column of Table~\ref{tab:compute}.

\begin{table}[h]
\centering
\footnotesize
\setlength{\tabcolsep}{4pt}
\begin{tabular}{@{}lccc@{}}
\toprule
\textbf{Method} & \textbf{Tokens/node} & \textbf{GNN (s)} & \textbf{Acc.\ (\%)} \\
\midrule
\llmgnn{} & $\sim$6k & $\sim$20 & $70.8$ \\
\locle{}         & $\sim$6k & $\sim$50 & $75.4$ \\
\rowcolor{ourrow}
\methodname{}    & $\sim$6k & $\sim$25 & $\mathbf{76.6}$ \\
\bottomrule
\end{tabular}
\caption{Cost vs.\ accuracy at matched budget.
\emph{Tokens/node}: LLM annotation tokens per node. \emph{GNN (s)}: training wall-clock per seed on GCN, excluding
annotation. \emph{Acc.}: mean accuracy.}
\label{tab:compute}
\vspace{-15pt}
\end{table}

\section{Related Work}
\label{sec:related}

\paragraph{Label-free node classification on TAGs.}
LLM annotation is a standard low-cost substitute for human labels~\citep{gilardi2023chatgpt}, first used as a GNN label source by \citet{chen2023exploring}.
Subsequent label-free TAG pipelines, such as \llmgnn{}~\citep{chen2024llmgnn}, \dma{}~\citep{sheng2025dma}, and \locle{}~\citep{zhang2025locle}, model LLM reliability at class granularity at best, treating it as uniform or class-conditional.
\methodname{} instead estimates a \emph{cluster-conditional} matrix and threads it through pseudo-label expansion and iterative correction.
A parallel line uses LLMs to \emph{enhance} features rather than annotate~\citep{he2024tape,zhao2023glem,wang2025tans,jin2024llmgraph}, orthogonal to and combinable with our annotation-side contribution.

\paragraph{Learning with noisy labels.}
Outside graphs~\citep{song2022noisy}, forward correction~\citep{patrini2017forward} calibrates the loss with a transition matrix, early-learning regularization~\citep{liu2020elr} curbs memorization, and clusterability~\citep{zhu2021clusterability} recovers a transition matrix from feature proximity without anchors, which we adapt for the first label-free per-cluster estimator.
Our cluster-conditional model sits between class-conditional and instance-dependent noise~\citep{xia2020part,cheng2022manifold,yao2021causalnl}; it is a structural special case of the instance-dependent graph noise benchmarked by \citet{begin2025}.
Extended discussion, with graph active learning and LLMs-on-graphs, is in Appendix~\ref{app:related}.

\section{Conclusion}
\label{sec:conclusion}
We showed, and empirically validated, that LLM annotation noise on graphs is
cluster-conditional rather than class-conditional, and proposed \methodname{}, a
label-free pipeline that estimates a per-cluster transition matrix and
threads it through pseudo-label expansion and iterative correction. Across various
benchmarks and GNN backbones, \methodname{} beats the previous state of the art baseline methods, with gains that track the cluster-conditional structure our
diagnostic measures. Cluster-conditional noise modeling is a natural next step
on the precision ladder beyond class-conditional approaches, and combining it with
orthogonal mechanisms such as graph rewiring is the clear way forward.

\newpage
\section*{Limitations}
\label{sec:limitations}

\methodname{} targets substantial, locally structured LLM noise, and its scope is bounded in three main respects. First, it does not modify graph structure; on small, dense, and
homophilous graphs such as \textsc{Cora}---where our estimator support
is also sparsest---methods utilizing structural rewiring (like
\locle{}) outperform it, and the deficit persists across annotators
(§\ref{sec:exp-annotator}), as a support-limited account predicts. Second, the cross-modality agreement estimator assumes partially independent neighbour errors. While the signal remains strongly discriminative overall, spatially coherent mislabeling can cause $T_c$ to over-read reliability in regions where neighbours share the same LLM annotation mistake. Finally, the approach requires sufficient probe support to estimate $T_c$, placing large, many-class graphs with weak feature-space structure (e.g., \textsc{ogbn-arxiv}) out of scope. In such sparse regimes---where the low silhouette score is an intrinsic graph property that budget increases cannot resolve---\methodname{} safely degrades to a class-conditional pipeline rather than failing outright. Practitioners can identify these cases a-priori using the label-free diagnostics of §\ref{sec:deployment}, while future work might address them by integrating structural rewiring, pooling related classes, or extending the reliability estimator to continuous feature spaces.
Relatedly, all five benchmarks are homophilous ($h = 0.66$--$0.81$);
whether cross-modality agreement remains a calibrated reliability signal
on heterophilous graphs, where graph and feature neighbours disagree by
construction, is untested and left to future work.

\section*{Ethical Considerations}
\label{sec:ethics}
\methodname{} replaces human labels with LLM annotations, so it inherits the risks that come with any LLM-generated supervision. LLM annotation errors are not uniform, and on graphs they concentrate in particular regions, which means a model trained on these labels can pick up and reinforce the annotator's biases in the places where the annotator is weakest. This is a fairness concern when those regions correspond to under-represented classes or communities. Our method reduces the problem by estimating where annotations are unreliable and trusting them less in those regions, but it does not remove the bias, and the reliability estimate is itself approximate. We therefore encourage practitioners to check label quality on their own data, particularly for minority groups, before acting on the predictions.

All of our experiments use public citation and web-graph benchmarks that contain no personal or sensitive information. We note, however, that the same pipeline applied to graphs of people, such as social or communication networks, could lower the cost of profiling or surveillance, and we discourage such uses. On the resource side, the method queries a hosted LLM on a small per-class budget and trains standard, small GNNs, so its computational and environmental cost is limited, and its aim is to cut annotation effort rather than to train larger models.

\section*{Acknowledgements}
The NSF partially supported this work under grants IIS-2528540, IIS-2533551, CNS-2426514, and CMMI-2146076.
Any opinions, findings, conclusions, or recommendations expressed in this material are those of the authors and do not necessarily reflect the views of the sponsors.

\bibliography{references}

\appendix

\section{Cluster-Conditional Noise Diagnostic: Full Setup and Extended
Findings}
\label{app:diag}

This appendix provides the full cluster-conditional noise diagnostic that motivates \methodname{} and whose headline is stated in the Introduction: how the clusters are formed, the three statistics we report, the construction of the synthetic null control, and the extended per-dataset evidence.

\subsection{Clustering and per-cluster accuracy}
\label{app:diag-setup}

For each dataset, we take the same LLM annotations that the
downstream pipeline uses (\S\ref{sec:experiments}), set $K = 2C$ for $C$
classes, and run $K$-means on the embeddings of a self-supervised graph
encoder (GraphMAE2~\citep{hou2023graphmae2}, pretrained on the full
unlabeled graph by masked-feature reconstruction). The choice $K\!=\!2C$
is the same one used by our method to estimate $T_c$
(\S\ref{sec:method}); the diagnostic is therefore a direct readout of
the noise model the method commits to. For every cluster $c$ and true
class $i$ we estimate
$T_c[c,i,i]$, the probability the LLM is correct on class-$i$ nodes
that fall in cluster $c$, using all (cluster, class) cells with at
least 20 ground-truth-labelled examples to avoid small-sample artefacts.

\subsection{Three statistics for shape}
\label{app:diag-stats}

We summarise per-cluster accuracy with three numbers:

\begin{itemize}[topsep=2pt,itemsep=2pt,leftmargin=*]
  \item $T_{ii}$, the global per-class LLM accuracy---the single number
        a class-conditional matrix would assign to a class. We report it
        as the per-class (macro) mean over classes; this differs from the
        class-frequency-weighted overall accuracy in
        Table~\ref{tab:datastats} on the class-imbalanced datasets
        (e.g.\ \textsc{PubMed}: $0.76$ macro vs.\ $88\%$ overall).
  \item $\bar\Delta$, the within-class accuracy gap (max minus min of
        $T_c[c,i,i]$ across clusters, averaged over classes).
        $\bar\Delta$ is an intuitive raw range, but it inflates when
        clusters carry little support; we treat it as a rough scale and
        rely on $\bar F$ for significance.
  \item $\bar F$, the one-way ANOVA F-statistic for per-cluster
        accuracy, averaged over classes. $\bar F$ asks whether
        per-cluster accuracy varies across clusters by more than the
        within-cluster sampling noise would explain, so a strictly
        class-conditional noise model predicts $\bar F \approx 1$.
\end{itemize}

The combined $p$ in Table~\ref{tab:diag} is Fisher's combined $p$-value
across per-class ANOVA tests.

\subsection{Diagnostic results}
\label{app:diag-results}

\begin{figure*}[t]
\centering
\includegraphics[width=\linewidth]{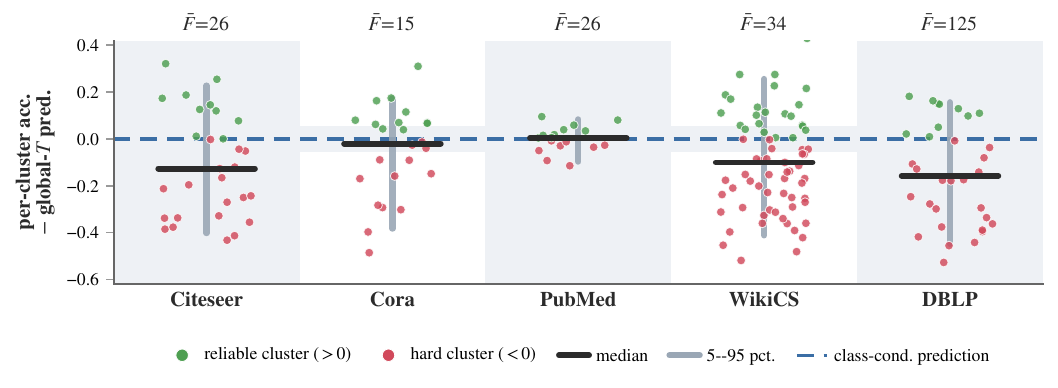}
\caption{Per-cell deviation of true per-cluster LLM accuracy from what
a global $T$ predicts ($T_c[k,i,i] - T[i,i]$; well-supported cells;
bars 5--95 percentile, ticks medians). A class-conditional model puts
every point at zero (red dashed line). The near-class-conditional
\textsc{PubMed} stays at zero; the other four benchmarks scatter by
tens of points and skew negative---a global correction
\emph{under}-corrects the hard clusters.}
\label{fig:diagnosis}
\end{figure*}

Table~\ref{tab:diag} reports the three statistics across the five
benchmarks. The class-conditional null is rejected on every one, with
$\bar F$ climbing from $14.8$ on \textsc{Cora} to $125$ on \textsc{DBLP}
(combined $p < 10^{-70}$), whereas the synthetic control of
\S\ref{app:diag-control} sits at $\bar F = 1.1$ ($p = 0.19$); the
structure therefore belongs to the LLM, not to the diagnostic.
Figure~\ref{fig:diagnosis} sharpens the point: a global matrix
systematically \emph{over}-predicts accuracy on the hardest clusters,
leaving a global correction to under-correct precisely the nodes that
need it most. \textsc{PubMed} is the lone near-class-conditional exception: although its
$\bar F = 25.7$ remains significant---it is by far the largest graph
($\sim\!20$K nodes), so even sub-percentage per-cluster differences clear
the F-test---its effect size $\bar\Delta = 0.14$ is several times smaller
than on the other benchmarks, leaving a per-cluster view little to add over
a global one. This later marks where \methodname{} does and does not help
(\S\ref{sec:experiments}).

\begin{table}[t]
\centering
\small
\begin{tabular}{@{}lcccc@{}}
\toprule
\textbf{Dataset} & $T_{ii}$ & $\bar\Delta$ & $\bar F$ & comb.\ $p$ \\
\midrule
\textsc{CiteSeer} & 0.65 & 0.71 & 25.8  & $<\!10^{-70}$ \\
\textsc{Cora}     & 0.59 & 0.55 & 14.8  & $<\!10^{-70}$ \\
\textsc{PubMed}   & 0.76 & 0.14 & 25.7  & $<\!10^{-70}$ \\
\textsc{WikiCS}   & 0.68 & 0.65 & 34.2  & $<\!10^{-70}$ \\
\textsc{DBLP}     & 0.71 & 0.62 & 125.3 & $<\!10^{-70}$ \\
\bottomrule
\end{tabular}
\caption{The class-conditional null is rejected on every benchmark
($\bar F \gg 1$, $p < 10^{-70}$). $T_{ii}$: global per-class LLM
accuracy. $\bar\Delta$: within-class max--min accuracy gap across
clusters (raw scale). Synthetic class-conditional null control returns
$\bar F = 1.1$ ($p = 0.19$), confirming the diagnostic itself does not
manufacture cluster structure (\S\ref{app:diag-control}).}
\label{tab:diag}
\end{table}

\subsection{Synthetic null control}
\label{app:diag-control}

A correct diagnostic must not invent cluster structure where none
exists. To check this, we construct a synthetic benchmark with
class-conditional noise \emph{by construction}: starting from a held-out
TAG that we do not otherwise use in the paper, we replace each node's
true label with a draw from a hand-set per-class confusion matrix
(diagonal $0.62$, off-diagonals uniform), so the noise is class-conditional
by construction at the population level. We then run the diagnostic
above. Because the noise is class-conditional, the F-test should not
reject the null regardless of how the nodes happen to cluster; any
$\bar F \gg 1$ would indicate a methodological artefact rather than
real noise structure. We observe $\bar F = 1.1$ ($p = 0.19$), confirming
that the diagnostic does not manufacture cluster structure.

\subsection{Extended findings}
\label{app:diag-findings}

\paragraph{A global matrix mispredicts per-cluster accuracy by tens of
points.} On \textsc{Cora}, \textsc{CiteSeer}, \textsc{WikiCS} and
\textsc{DBLP} the gap between true per-cluster accuracy and the global
prediction routinely exceeds $\pm 0.2$ and reaches $\pm 0.4$. On
\textsc{DBLP} a single true class is annotated at $21.6\%$ accuracy in
its hardest cluster and $81.0\%$ in its easiest, against a global
prediction of $64.7\%$. The gap is also \emph{skewed}: a global $T$
systematically over-predicts accuracy on the hard clusters, so a global
correction under-corrects exactly the nodes that need it most.

\begin{figure*}[t]
\centering
\includegraphics[width=\textwidth]{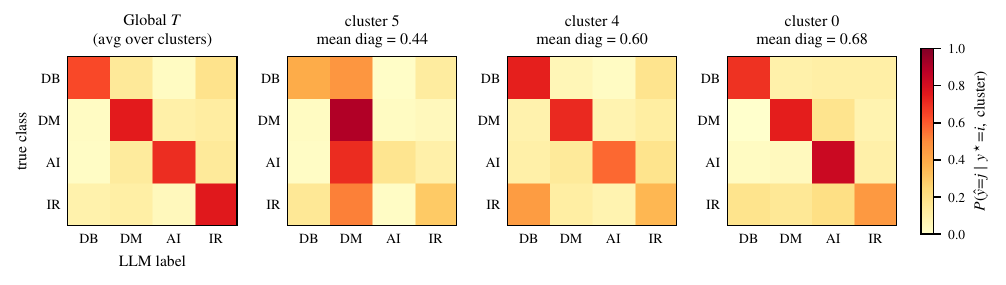}
\caption{\textbf{Extended diagnostic.} \textsc{DBLP}: the global matrix
$T$ is a blurred average of three structurally different per-cluster
matrices $T_c$.}
\label{fig:dblp_heatmap}
\end{figure*}

\paragraph{A global matrix is the average of structurally different
matrices.} Figure~\ref{fig:dblp_heatmap} illustrates this on
\textsc{DBLP}: three per-cluster matrices that differ not only in their
diagonal level but in \emph{which} off-diagonal confusions dominate;
the global $T$ is a blurred average that fits none of them.

\paragraph{The magnitude varies by dataset.} \textsc{DBLP} and
\textsc{WikiCS} have the strongest cluster structure ($\bar F = 125$
and $34$); \textsc{Cora} the weakest among the five ($\bar F = 14.8$).
This variation predicts where our correction helps most
(\S\ref{sec:experiments}).

\paragraph{The clusterability assumption holds empirically.} The
label-free estimator of Eq.~\ref{eq:tc-agreement} rests on one
assumption: that clusters are semantically coherent, so that
neighbour agreement within a cluster reflects the reliability of its
dominant true class. We verify both halves directly. The GraphMAE2 clusters are coherent: their
size-weighted single-class purity is $63$--$80\%$ across the five
benchmarks (well above the $1/C$ chance level), and the mode is a good
proxy: the most frequent LLM label coincides with the cluster's dominant
\emph{true} class in $83$--$88\%$ of clusters on \textsc{PubMed},
\textsc{WikiCS} and \textsc{DBLP}, and in $75$--$79\%$ on
\textsc{CiteSeer} and \textsc{Cora} (see Table~\ref{tab:clusterability}). The
shrink-to-uniform back-off for low-purity clusters
(\S\ref{sec:method-estimate}) is precisely what guards the minority of
clusters where the proxy is unreliable.

\section{Extended Related Work}
\label{app:related}

This appendix expands the condensed discussion of Section~\ref{sec:related}.

\paragraph{Label-free node classification on TAGs.}
LLMs now annotate data at near-human quality across many NLP tasks~\citep{gilardi2023chatgpt,ding2023gpt3annotator}, which \citet{chen2023exploring} first turned into low-budget supervision for GNNs.
The label-free pipelines that followed differ mainly in how they treat the resulting noise: \llmgnn{} assumes it is roughly uniform and relies on confidence-aware selection and post-filtering~\citep{chen2024llmgnn}; \dma{} estimates a class-conditional confusion matrix from synthetic probes and applies forward correction~\citep{sheng2025dma,patrini2017forward}; and \locle{} forgoes an explicit noise model for a five-stage iterative refinement (subspace selection, consistency filtering, re-annotation, rewiring, rank-and-correct)~\citep{zhang2025locle}. \methodname{} replaces the class-level noise model these systems use, explicitly or implicitly, with a cluster-conditional one and threads it through every stage.
Reliability estimation also extends beyond annotation: \citet{huang2026counterfactual} discounts agreement where communication correlates agent errors, and \citet{huang2026evolverouter} models prompt-dependent accuracy.

\paragraph{Graph active learning.}
Since the budget is small relative to the graph, \emph{which} nodes to label is a first-order decision. Classical selectors mix structural signals (centrality, feature propagation) with model uncertainty and learned  policies~\citep{cai2017active,wu2019featprop,hu2020gpa}, while later work targets the noisy and label-free regimes via reliability- and influence-weighted, partition-based, and confidence-filtered selection~\citep{zhang2021rim,ma2022partition,chen2024llmgnn}.
We keep \locle{}'s subspace-clustering selection~\citep{zhang2025locle} as the representativeness backbone (\S\ref{sec:method-annotate}); reliability is estimated locally only \emph{after} annotation and is applied downstream rather than to candidate scoring, since no labels exist before querying.
Budgeted selection over a graph is not specific to node
labelling; \citet{li2025mass} ranks pretraining corpora through
a skill graph.

\paragraph{Learning with noisy labels.}
Outside graphs, training under label noise is well studied~\citep{song2022noisy}: \methodname{} reuses two standard tools, forward correction~\citep{patrini2017forward} and early-learning regularisation~\citep{liu2020elr}, and adapts clusterability~\citep{zhu2021clusterability}---which recovers a transition matrix from feature proximity without anchor points---to obtain the first label-free per-cluster estimator.
Between class-conditional and fully instance-dependent noise~\citep{xia2020part,cheng2022manifold,yao2021causalnl}, our cluster-conditional model is the special case in which the instance dependence (benchmarked on graphs by \citet{begin2025}) is organised by feature-space clusters.

\paragraph{LLMs on graphs: enhancers and predictors.}
Beyond annotation, LLMs are coupled with graph learning either as \emph{enhancers}, enriching node features for a downstream GNN through graph-aware encoders, free-text explanations, joint training, or synthesised descriptions~\citep{chien2022giant,he2024tape,zhao2023glem,wang2025tans, thapaliya2025semantic}, or as \emph{predictors}, answering directly from a verbalised graph via instruction tuning, token-space projection, or cross-domain
pretraining~\citep{ye2024instructglm,tang2024graphgpt,chen2024llaga,liu2024ofa}; \citet{jin2024llmgraph} survey both. \methodname{} works in the annotator regime and is orthogonal to these: a future combination could supply the embeddings it clusters on with LLM-enhanced features, or extend the cluster-conditional estimator to heterogeneous graphs with typed, multimodal node content~\citep{zhang2019heterogeneous} and to settings where graph structure is shared across data modalities and tasks~\citep{li2026graphsubstrate}.

\section{Iterative Correction Convergence}
\label{app:ilc-conv}

The $T_c$-gated iterative label correction of \S\ref{sec:method-ilc}
retrains the GNN and revises the label pool until no label changes.
Table~\ref{tab:ilc-conv} reports its behaviour across all ten cells,
read from the per-round correction logs of the main \methodname{} runs
(5-seed means; no extra experiment). Three properties hold uniformly. \emph{(i) It converges quickly}: every cell stops within $2.4$--$4.0$ rounds.
\emph{(ii) Corrections decay monotonically}: the first round accounts for $83$--$98\%$ of all label changes, the second round for a handful, and rounds three onward for almost none, there is no oscillation, and the ``no label changes'' stopping rule is reached fast.
\emph{(iii)
Correction volume tracks dataset noise}: \textsc{PubMed}, whose LLM
labels are the cleanest ($\approx\!88\%$ accurate), changes the fewest
labels ($\approx\!12$), whereas the noisier \textsc{CiteSeer} and
\textsc{WikiCS} change the most ($\approx\!48$--$52$), consistent with
the noise diagnostic of Appendix~\ref{app:diag}.

\begin{table}[h]
\centering
\small
\setlength{\tabcolsep}{4pt}
\begin{tabular}{@{}lccccc@{}}
\toprule
\textbf{Metric} & \textbf{Cite.} & \textbf{Cora} & \textbf{PubM.} & \textbf{Wiki} & \textbf{DBLP} \\
\midrule
\rowcolor{grouphdr}\multicolumn{6}{@{}l}{\textit{GAT}}\\
Rounds         & 4.0  & 3.0  & 2.4  & 3.4  & 3.6  \\
Labels corr.   & 41.0 & 29.4 & 12.4 & 37.2 & 36.2 \\
\% in round 1  & 83   & 95   & 97   & 85   & 91   \\
\midrule
\rowcolor{grouphdr}\multicolumn{6}{@{}l}{\textit{GCN}}\\
Rounds         & 3.8  & 3.0  & 2.4  & 3.2  & 2.6  \\
Labels corr.   & 47.8 & 28.8 & 12.2 & 52.0 & 33.8 \\
\% in round 1  & 87   & 94   & 97   & 95   & 98   \\
\bottomrule
\end{tabular}
\caption{Convergence of $T_c$-ILC (5-seed means). The correction loop
stops in at most four rounds on every cell, the first round accounts for
$83$--$98\%$ of all corrections, and the number of labels corrected
tracks dataset noise (fewest on the cleanest dataset, \textsc{PubMed}).}
\label{tab:ilc-conv}
\end{table}

\section{Per-Dataset Budget Sensitivity}
\label{app:budget}

Table~\ref{tab:budget-full} gives the per-dataset numbers behind the
aggregate budget curve of Figure~\ref{fig:budget}. The full budget is
$B=50C$ ($50$ queries per class for a $C$-class graph); the
$25/50/75/100\%$ levels in the table set the per-class budget to $13$,
$25$, $37$, and $50$ queries respectively. At each level we re-select and
re-annotate from scratch and run both methods with all other settings
fixed, on GCN over $5$ seeds.

\begin{table*}[t]
\centering
\small
\setlength{\tabcolsep}{6pt}
\begin{tabular}{@{}lccccc@{}}
\toprule
\textbf{Method} & \textbf{CiteSeer} & \textbf{Cora} & \textbf{PubMed} & \textbf{WikiCS} & \textbf{DBLP} \\
\midrule
\rowcolor{grouphdr}\multicolumn{6}{@{}l}{\textit{25\% budget}}\\
\locle{}       & \val{69.30}{2.02} & \valb{72.73}{1.41} & \val{68.69}{3.66} & \val{64.01}{2.30} & \val{62.43}{3.42} \\
\rowcolor{ourrow}
\methodname{}  & \valb{73.45}{1.31} & \val{67.16}{3.08} & \valb{79.59}{1.78} & \valb{71.07}{2.50} & \valb{69.36}{3.79} \\
\midrule
\rowcolor{grouphdr}\multicolumn{6}{@{}l}{\textit{50\% budget}}\\
\locle{}       & \val{73.78}{0.33} & \valb{75.31}{1.29} & \val{77.30}{1.17} & \val{69.06}{0.13} & \val{62.59}{4.70} \\
\rowcolor{ourrow}
\methodname{}  & \valb{74.81}{0.55} & \val{74.32}{0.27} & \valb{81.50}{0.70} & \valb{71.26}{1.48} & \valb{74.23}{1.93} \\
\midrule
\rowcolor{grouphdr}\multicolumn{6}{@{}l}{\textit{75\% budget}}\\
\locle{}       & \val{74.48}{0.59} & \valb{77.17}{0.89} & \val{80.77}{0.52} & \valb{74.11}{0.38} & \val{72.21}{1.83} \\
\rowcolor{ourrow}
\methodname{}  & \valb{75.82}{0.82} & \val{74.40}{1.01} & \valb{82.00}{0.39} & \val{73.30}{0.47} & \valb{76.17}{1.05} \\
\midrule
\rowcolor{grouphdr}\multicolumn{6}{@{}l}{\textit{100\% budget}}\\
\locle{}       & \val{73.62}{0.87} & \valb{79.81}{0.57} & \val{81.69}{0.85} & \valb{72.73}{0.82} & \val{72.89}{2.47} \\
\rowcolor{ourrow}
\methodname{}  & \valb{75.92}{0.53} & \val{75.04}{0.48} & \valb{81.85}{0.65} & \val{72.70}{2.23} & \valb{75.66}{0.86} \\
\bottomrule
\end{tabular}
\caption{Per-dataset budget sensitivity (test accuracy $\%$, GCN,
5-seed mean${}_{\pm\text{std}}$) at $25/50/75/100\%$ of each method's
full annotation budget. \textbf{Bold} marks the better of the two
methods in each cell. The aggregate is plotted in
Figure~\ref{fig:budget}.}
\label{tab:budget-full}
\end{table*}

\section{Ablations and Additional Analyses}
\label{app:ablation}

Table~\ref{tab:ablation} reports two ablations, each changing a
\emph{single} setting of the locked \methodname{} pipeline and re-running
it for $3$ seeds; all other hyperparameters, the annotation cache, and the
budget are held at their main-experiment values. \textbf{(a)~ELR:} we
switch off the early-learning-regularization term
(ELR;~\citep{liu2020elr}) and change nothing else. Among the noise-robust
training components it is the load-bearing one on GCN---removing it costs
$0.34$pp on average, carrying the noisier text graphs (\textsc{CiteSeer}
$-1.95$, \textsc{PubMed} $-1.22$, \textsc{DBLP} $-1.30$); on GAT it is
roughly neutral, and \textsc{WikiCS} prefers no ELR on both backbones,
consistent with its lower label noise. \textbf{(b)~Clustering:} we replace
the default partition---$K\!=\!2C$ $K$-means on the GraphMAE2
embeddings---with $K$-means on raw node features (all five datasets) or
spectral clustering of the same embeddings (run on the two representative
graphs, sparse \textsc{Cora} and dense \textsc{WikiCS}), and re-estimate
$T_c$ on each new partition before re-running the pipeline. Accuracy moves
by at most $0.8$pp, so the method is insensitive to the partitioner---the
cluster-conditional effect is a property of the data, not of one
clustering choice (cf.\ the null control of
Appendix~\ref{app:diag-control}).

\begin{table*}[h]
\centering
\small
\setlength{\tabcolsep}{7pt}
\renewcommand{\arraystretch}{1.0}
\begin{tabular}{@{}lccccc@{}}
\toprule
 & \textsc{CiteSeer} & \textsc{Cora} & \textsc{PubMed} & \textsc{WikiCS} & \textsc{DBLP} \\
\midrule
\rowcolor{grouphdr}\multicolumn{6}{@{}l}{\textit{(a) Early-learning regularization ($\Delta=$ $-$ELR $-$ full \methodname{})}}\\
\methodname{} (GAT)  & \val{75.09}{0.77} & \val{75.68}{0.47} & \val{80.87}{0.69} & \val{72.84}{2.70} & \val{75.42}{3.15} \\
\;\;$-$ELR (GAT)     & \val{74.80}{0.69} & \val{75.04}{0.55} & \val{81.50}{0.83} & \val{75.99}{2.03} & \val{75.09}{1.24} \\
\;\;$\Delta$         & $-0.29$ & $-0.64$ & $+0.63$ & $+3.15$ & $-0.33$ \\
\methodname{} (GCN)  & \val{76.27}{0.30} & \val{75.30}{0.35} & \val{82.36}{0.67} & \val{73.67}{0.30} & \val{75.86}{0.64} \\
\;\;$-$ELR (GCN)     & \val{74.33}{0.94} & \val{75.23}{0.94} & \val{81.14}{0.83} & \val{76.49}{2.02} & \val{74.56}{1.97} \\
\;\;$\Delta$         & $-1.95$ & $-0.06$ & $-1.22$ & $+2.83$ & $-1.30$ \\
\midrule
\rowcolor{grouphdr}\multicolumn{6}{@{}l}{\textit{(b) Clustering procedure (GCN)}}\\
GraphMAE2 ($K$-means)  & \val{76.27}{0.30} & \val{75.30}{0.35} & \val{82.36}{0.67} & \val{73.67}{0.30} & \val{75.86}{0.64} \\
Raw feat.\ ($K$-means) & \val{76.22}{0.59} & \val{75.37}{0.15} & \val{81.79}{0.15} & \val{74.50}{0.86} & \val{76.00}{0.44} \\
Spectral (GraphMAE2)   & \na & \val{75.65}{0.25} & \na & \val{73.84}{0.20} & \na \\
\bottomrule
\end{tabular}
\caption{\textbf{Ablations} (test accuracy $\%$, 3-seed
mean${}_{\pm\text{std}}$). \textbf{(a)} Removing early-learning
regularization ($\Delta$ negative $=$ ELR helps; $\Delta$ on means).
\textbf{(b)} Swapping the clustering procedure (GCN; spectral on the two
representative graphs).}
\label{tab:ablation}
\end{table*}

\subsection{Stability of $T_c$ Under Re-clustering}
\label{app:stability}

The estimator consumes a $K$-means partition, and $K$-means depends on
initialisation. We re-ran the label-free estimation under five random
clustering seeds with the annotation cache and every hyperparameter held
fixed. The estimate is stable: across \textsc{CiteSeer}, \textsc{PubMed},
\textsc{WikiCS}, and \textsc{DBLP}, the dataset-level mean diagonal
varies by a standard deviation of at most $0.002$ and per-node
reliability by at most $0.004$. \textsc{Cora} is the exception (std
$0.021$, an order of magnitude larger), consistently with the rest of
our analysis: it is the most fragmented setting, with $28.6\%$ of cells
backing off, so its estimate is the one most exposed to how the
partition happens to fall. The
insensitivity to $K$ (Figure~\ref{fig:ksens}) and to the clustering
procedure (Table~\ref{tab:ablation}) has the same explanation as here:
each cell's reliability is estimated from neighbour agreement over all
nodes, not from the few probe labels that land inside it.

\subsection{Behaviour on Low-Purity Clusters}
\label{app:lowpurity}

Clusters mixing several true classes are where a concentration-based
estimate would fail, and what the shrink-to-uniform back-off is designed
for. Two cases show both directions of the intended behaviour.
\textsc{WikiCS} cluster 16 (388 nodes, $43\%$ single-class pure)
receives a low self-trust of $T_c[16,9,9]=0.31$ with the remaining mass
spread over five classes, so expansion into it tightens and its labels
are revisited more often. \textsc{DBLP} cluster 4 is comparably mixed
($52\%$ pure), yet the estimate keeps class-2 trust high at $0.74$, so
reliable labels inside a noisy region are retained. The second case is
the one a purity-based estimator would get wrong, and it is why the
estimate conditions on cluster and class jointly rather than scoring
clusters as a whole.

\subsection{Accuracy of the Correction Step}
\label{app:correction-accuracy}

Appendix~\ref{app:ilc-conv} reports \emph{how many} labels the
$T_c$-gated correction loop changes; this section reports whether those
changes are right. We instrumented the loop to log the old label, the
new label, and the ground-truth label for every correction across all
ten (dataset, backbone) settings and 5 seeds. Ground truth is read only
for this measurement and never enters the pipeline. Because a correction
always moves a label to a different class, the relevant chance baseline
is $1/(C-1)$, i.e.\ $50\%$ on \textsc{PubMed} ($C{=}3$) down to $11\%$
on \textsc{WikiCS} ($C{=}10$).

Of the $4{,}787$ corrections logged (counted per label change per round
across the full label pool, so the total exceeds the per-run figures of
Table~\ref{tab:ilc-conv}), $56.8\%$ move the label to the true class and $23.0\%$ overwrite a label that was already correct; the
remainder replace one wrong label with another. The step is therefore
net-positive with respect to ground-truth accuracy in nine of the ten
settings, with an overall net gain of $+1{,}617$ correct labels.
Per-setting to-true rates range from $50\%$ to $88\%$, with the highest
rate on \textsc{Cora}-GCN ($87.5\%$). The rate is not a proxy for
estimator quality: \textsc{Cora} is our sparsest estimate
(Appendix~\ref{app:error-analysis}) but also our most homophilous graph
($h=0.81$), so the neighbour-support requirement of
\S\ref{sec:method-ilc} is met mainly by genuinely well-connected
same-class regions, and the loop changes comparatively few labels
(Table~\ref{tab:ilc-conv}). The step is therefore net-positive even on
the setting where cluster-conditional gating adds least.
\section{Clusterability: Quantitative Validation}
\label{app:clusterability}

The label-free $T_c$ estimator rests on two
assumptions: that GraphMAE2 clusters are class-coherent, and that the
most frequent LLM label in a cluster proxies its dominant true class.
Table~\ref{tab:clusterability} validates both. All three quantities are
measured on the same $K\!=\!2C$ GraphMAE2 partition the method uses, with
ground-truth labels read only to score them (the estimator never sees
them). Clusters are $63$--$80\%$ single-class pure on average, a majority
are predominantly one class, and the LLM mode equals the dominant true
class in $75$--$88\%$ of clusters, confirming that the GraphMAE2
partition carries genuine class structure for the estimator to exploit.

\begin{table}[h]
\centering
\small
\setlength{\tabcolsep}{5pt}
\renewcommand{\arraystretch}{1.0}
\begin{tabular}{@{}lcccc@{}}
\toprule
\textbf{Dataset} & $K$ & \textbf{Purity} & \textbf{\%Maj.} & \textbf{Mode$=$T} \\
\midrule
\textsc{CiteSeer} & 12 & 0.70 & 92\% & 75\% \\
\textsc{Cora}     & 14 & 0.80 & 86\% & 79\% \\
\textsc{PubMed}   & 6  & 0.63 & 83\% & 83\% \\
\textsc{WikiCS}   & 20 & 0.70 & 75\% & 85\% \\
\textsc{DBLP}     & 8  & 0.78 & 100\% & 88\% \\
\bottomrule
\end{tabular}
\caption{\textbf{Clusterability validation.} Per-cluster mean
dominant-true-class fraction (Purity); fraction of clusters that are
majority single-class (\%Maj.); fraction whose most-frequent LLM label
equals the dominant true class (Mode$=$T). Ground truth used for
measurement only.}
\label{tab:clusterability}
\end{table}

\section{Experimental Setup Details}
\label{app:setup}

This appendix expands the condensed setup of \S\ref{sec:exp-setup}.

\begin{table}[t]
\centering
\small
\setlength{\tabcolsep}{3.5pt}
\begin{tabular}{@{}lrrrrcc@{}}
\toprule
\textbf{Dataset} & \textbf{\#Nodes} & \textbf{\#Edges} & \textbf{\#Cls} & \textbf{Deg.} & \textbf{Hom.} & \textbf{LLM} \\
\midrule
\textsc{CiteSeer} & $3{,}186$  & $4{,}277$   & 6  & 2.7  & 0.79 & 65 \\
\textsc{Cora}     & $2{,}708$  & $5{,}429$   & 7  & 4.0  & 0.81 & 68 \\
\textsc{PubMed}   & $19{,}717$ & $44{,}335$  & 3  & 4.5  & 0.80 & 88 \\
\textsc{WikiCS}   & $11{,}701$ & $215{,}863$ & 10 & 36.9 & 0.66 & 67 \\
\textsc{DBLP}     & $14{,}376$ & $215{,}663$ & 4  & 30.0 & 0.67 & 69 \\
\bottomrule
\end{tabular}
\caption{Dataset statistics: number of nodes, edges, classes (\#Cls),
average degree (Deg.), edge homophily (Hom.), and raw \textsc{gpt-3.5}
LLM annotation accuracy (LLM, \%).}
\label{tab:datastats}
\end{table}

\paragraph{LLM annotator.}
The annotator is \texttt{gpt-3.5-turbo}, queried with $n\!=\!3$ self-consistency and chain-of-thought prompting under the same prompt template used by \locle{} and \llmgnn{}~\citep{chen2024llmgnn}.
At $\sim\!2{,}000$ tokens per query, the per-query token cost is identical across all three pipelines.

\subsection{Annotation Prompt}
\label{app:prompt}

For full reproducibility we use the same zero-shot annotation prompt as
\locle{}~\citep{zhang2025locle} and \llmgnn{}~\citep{chen2024llmgnn};
\methodname{} introduces no changes to the annotation prompt. The template
(Figure~\ref{fig:prompt}) prepends a fixed task instruction, lists the
dataset's category names, supplies the target node's text (paper
title$+$abstract for the citation graphs, article text for WikiCS), and
requests a single answer with a self-reported confidence in a fixed
parseable format. Each node is queried with chain-of-thought reasoning and
$n\!=\!3$ self-consistency sampling, and the majority-voted answer is taken
as the LLM label, matching \locle{} exactly. The category list is the only
per-dataset component; we show the Cora instantiation in
Figure~\ref{fig:prompt} and substitute each dataset's own label set
otherwise.

\begin{figure*}[h]
\centering
\begin{promptbox}[Label-Free Annotation Prompt]
\small
\textbf{Task instruction:}\\
You are a model that is especially good at classifying a paper's category.
I will first give you all the possible categories and their explanation.
Please answer the following question: \emph{What is the category of this
paper?}

\medskip
\textbf{Categories} (\emph{Cora} shown; replaced per dataset by its own label set):\\
\texttt{[rule\_learning, neural\_networks, case\_based, genetic\_algorithms,
theory, reinforcement\_learning, probabilistic\_methods]}

\medskip
\textbf{Target paper:}\\
\{\textit{title and abstract of the node}\}

\medskip
\textbf{Answer format:}\\
Analyze the question step by step. Output your answer together with a
confidence ranging from $0$ to $100$, as a single-element list of Python
dicts, and output only the one answer you think is most likely:\\[2pt]
\ansslot{\texttt{[\{"answer": <answer>, "confidence": <confidence>\}]}}

\medskip
\textbf{Sampling:} queried with chain-of-thought reasoning and $n\!=\!3$
self-consistency; the majority-voted label is kept.
\end{promptbox}
\caption{Zero-shot LLM annotation prompt used to obtain node pseudo-labels.
\methodname{} reuses the \locle{}/\llmgnn{} template unchanged; only the
category list varies across datasets. The boxed line is the required
machine-parseable answer format.}
\label{fig:prompt}
\end{figure*}
\paragraph{Annotation budget.}
Each dataset's annotation budget matches \locle{}'s released per-dataset
budget.
Because
both the annotator and the budget match, total token usage at matched
budget is the same for \methodname{} and the baselines.

\paragraph{Preprocessing.}
We adopt \locle{}'s preprocessing unchanged, including its SentenceBERT
node features and graph construction, so the inputs are identical across
the compared pipelines.

\paragraph{Hyperparameter selection.}
\methodname{}'s single tunable knob, the self-training gate strength
$\alpha$ of Eq.~\ref{eq:tc-gate}, is selected by 3-seed cross-validation
per (backbone, dataset) on a held-out subset of the LLM annotation cache.
All remaining hyperparameters are dataset-agnostic and listed in
Appendix~\ref{app:hp}.

\section{Implementation Details}
\label{app:hp}

Beyond the single gate strength $\alpha$ (Eq.~\ref{eq:tc-gate}),
\methodname{} uses one fixed configuration shared across all five datasets
and both backbones, with no per-dataset or per-backbone tuning.

\paragraph{GNN backbone.} For a controlled comparison we keep the
backbone identical to our baselines~\citep{chen2024llmgnn,zhang2025locle}:
a two-layer, $64$-hidden-unit GCN or GAT trained for $300$ epochs with
Adam (learning rate $3\!\times\!10^{-3}$, weight decay
$5\!\times\!10^{-4}$, dropout $0.7$). These are the published baseline
settings rather than values we tuned, so the backbone is the same for all
compared methods.

\paragraph{Protocol.} Following the LoCLE annotation protocol, the budget
is $B=50C$ queries for a $C$-class graph, and the noise estimator
partitions each graph into $K=2C$ GraphMAE2~\citep{hou2023graphmae2}
clusters, a parameter-free rule tied to the class count rather than a
tuned value.

\paragraph{Thresholds of Eq.~\ref{eq:tc-gate}}
The base admission threshold is not tuned. It is set by a fixed,
dataset-agnostic rule from the probe-estimated per-class accuracy
$acc_c$, as $\tau_c = \mathrm{clip}(1 - 0.7 \cdot acc_c,\, 0.3,\, 0.7)$,
defaulting to $0.5$ for any class with no probe estimate. The same rule
applies on every dataset and backbone: a class the annotator handles
well admits pseudo-labels at a lower threshold, a class it handles
poorly at a higher one, and the clip bounds the range in both
directions. The gate strength $\alpha$ is \methodname{}'s only tuned
hyperparameter, selected by 3-seed cross-validation per (backbone,
dataset) on a held-out subset of the annotation cache. The correction
loop's $\theta_0$ and $\beta$ are fixed across all settings.

\begin{table}[h]
\centering
\small
\setlength{\tabcolsep}{5pt}
\begin{tabular}{@{}l>{\raggedright\arraybackslash}p{2.4cm}c@{}}
\toprule
\textbf{Symbol} & \textbf{Role} & \textbf{Value} \\
\midrule
$B$        & annotation budget        & $50C$ \\
$K$        & clusters                 & $2C$ \\
$\rho$     & probe fraction           & $0.4$ \\
$\tau_c$   & base admission threshold & $\mathrm{clip}(1{-}0.7\,acc_c,\,0.3,\,0.7)$ \\
$\alpha$ & gate strength (Eq.~\ref{eq:tc-gate}) & $0.3$ (typical; CV per cell) \\$\theta_0$ & base correction threshold & $0.5$ \\
$\beta$    & correction gate strength & $0.5$ \\
\bottomrule
\end{tabular}
\caption{Complete hyperparameter configuration. All values are fixed
across datasets and backbones except $\alpha$, selected by 3-seed
cross-validation per (backbone, dataset).}
\label{tab:hyperparams}
\vspace{-10pt}
\end{table}

\paragraph{Noise-robust components.} The robust-training and
label-denoising modules (early-learning regularization~\citep{liu2020elr},
edge dropout, GMM-based loss filtering, and the neighbour-agreement
correction step) are run with the default settings from their original
formulations; we adopt them unchanged. The expansion and correction loops
use conservative round-number acceptance thresholds and a fixed number of
iterations shared across all datasets, not selected per benchmark. The
ablations in Appendix~\ref{app:ablation} confirm the method is insensitive
to these choices, and the only dataset-specific input is the label-free
$T_c$ matrix estimated from each graph.

\section{When and Why \methodname{} Helps: An Estimator-Coverage Analysis}
\label{app:error-analysis}

To understand why \methodname{} helps on some (dataset, backbone) cells and not
others, we examine the estimated noise model itself. For each dataset we compare
the deployed transition tensor $T_c \in [0,1]^{K\times C\times C}$ (its row
$T_c[k,i,\cdot]$ is the estimated distribution of LLM labels for true class $i$
in cluster $k$) with the GraphMAE2 cluster assignments and the ground-truth
labels; the latter are used only to evaluate the estimate, never by the method.
Table~\ref{tab:casestudy} relates the quality of this estimate to the per-cell
gain over \locle{}.

\paragraph{Setup.} This is a post-hoc analysis of the \emph{deployed}
artifacts from the main experiment, not a new run. We use the same
$T_c$ that produces the results in Table~\ref{tab:main}: the
label-free cross-modality agreement estimator of Eq.~\ref{eq:tc-agreement},
fit once per dataset on the \textsc{gpt-3.5-turbo} annotations of the
active-selected nodes, over the $K=2C$ GraphMAE2 clusters (the same
partition used at training time). No
ground-truth labels enter $T_c$; they are loaded only to
\emph{measure} the quantities below, exactly as in
Appendix~\ref{app:clusterability}. We report four measurements per
dataset. (i)~\emph{LLM acc.}: accuracy of the raw LLM annotations against
ground truth (Table~\ref{tab:datastats}). (ii)~\emph{low-support}: the fraction
of the $K\!\times\!C$ (cluster, class) cells with too little probe evidence to
estimate cell-specific reliability, so they fall back to the cluster- and then
global-mean agreement (\S\ref{sec:method-estimate}) and carry no
\emph{cell-specific} signal. (iii)~\emph{$r$}: the
Pearson correlation, taken across clusters with at least five nodes,
between the estimated self-trust $\tfrac{1}{C}\sum_iT_c[k,i,i]$
and the ground-truth cluster purity (dominant-true-class fraction).
(iv)~\emph{$\Delta_{\text{GCN}}/\Delta_{\text{GAT}}$}: the $5$-seed-mean
accuracy gain over \locle{} from Table~\ref{tab:main}. The
cluster-level examples cited below are read directly off the deployed
$T_c$ and the GraphMAE2 partition, with cluster sizes and class
compositions measured from ground truth.

\begin{table}[t]
\centering
\small
\setlength{\tabcolsep}{4pt}
\renewcommand{\arraystretch}{1.0}
\begin{tabular}{@{}lccccc@{}}
\toprule
\textbf{Dataset} & \textbf{LLM} & \textbf{low-supp.} & \textbf{$r$} & \textbf{$\Delta_{\text{GCN}}$} & \textbf{$\Delta_{\text{GAT}}$} \\
 & acc.\% & \% & & & \\
\midrule
\textsc{CiteSeer} & 65 &  0.0 & 0.73 & $+2.30$ & $+2.95$ \\
\textsc{Cora}     & 68 & 28.6 & 0.41 & $-4.77$ & $+0.37$ \\
\textsc{PubMed}   & 88 & 33.3 & 0.70 & $+0.16$ & $+0.74$ \\
\textsc{WikiCS}   & 67 &  0.0 & 0.45 & $-0.03$ & $+6.05$ \\
\textsc{DBLP}     & 69 &  0.0 & 0.43 & $+2.77$ & $+0.92$ \\
\bottomrule
\end{tabular}
\caption{\textbf{Estimator coverage vs.\ per-cell gain.} Columns: raw LLM
accuracy; low-support; per-cluster calibration $r$; and gain over \locle{} by
backbone. Definitions in the text.}
\label{tab:casestudy}
\vspace{-12pt}
\end{table}

\paragraph{Where the estimate is informative.} The gains are largest on
\textsc{CiteSeer}, \textsc{WikiCS}, and \textsc{DBLP} (up to $+6.05$), and on
these datasets the estimate is both well-supported and well-calibrated: every
cluster$\times$class cell carries direct probe evidence ($0\%$ low-support), and
the estimated self-trust correlates with true cluster cleanliness
($r=0.73/0.45/0.43$, respectively). The individual cells also behave as intended.
In \textsc{WikiCS}, cluster $16$ ($388$ nodes, dominant true class $9$, only
$43\%$ pure) is assigned a low self-trust of $T_c[16,9,9]=0.31$ with the
remaining probability spread over five classes, so expansion into this region is
tightened and its labels are revisited more often. In \textsc{CiteSeer}, cluster $4$ ($206$ nodes) receives a low self-trust
of $T_c[4,0,0]=0.20$, which the ground-truth composition corroborates:
$109$ class-$0$ and $56$ class-$4$ nodes lie together, so the annotator
is unreliable for class $0$ in exactly this region.
\textsc{DBLP} cluster $4$
shows the value of conditioning on cluster and class jointly: although the
cluster is mixed ($52\%$ pure), the estimate keeps class-$2$ trust high ($0.74$),
so the reliable labels within it are retained even where the surrounding region
is noisy.

\paragraph{Where the estimate is too sparse (\textsc{Cora}).} The single clear
loss, \textsc{Cora}-GCN ($-4.77$), coincides with the sparsest estimate.
\textsc{Cora} is the most fragmented of the five settings: the smallest graph is
divided into the finest grid ($K\!=\!14$ clusters $\times\,C\!=\!7$ classes)
under the smallest budget ($50C\!=\!350$ queries), so many cells receive little
or no probe evidence: $28.6\%$ carry no cell-specific signal and revert to the
coarse mean, and over half rest on fewer than three annotations. Cluster $11$
($281$ nodes) is representative, as its class-$4$ cell draws no probe annotations
and falls back to the cluster mean, adding nothing beyond a global estimate. With
roughly a quarter of the noise model uninformative, \methodname{} has little
local structure to exploit, whereas \locle{}'s iterative confident learning and
propagation are well suited to a graph this small and homophilous ($h=0.81$). The
shortfall is specific to GCN; under GAT, whose attention discounts inconsistent
edges, the same configuration recovers to $+0.37$.

\paragraph{Where the estimate has little to add (\textsc{PubMed}).}
\textsc{PubMed} sits at the opposite extreme. Its LLM labels are already $88\%$
accurate, yet the estimate is both sparse ($33\%$ low-support) and conservative:
its informative rows carry the lowest mean self-trust of any dataset ($0.575$),
so it somewhat over-states the noise on an essentially clean graph. \methodname{}
responds cautiously, correcting only about $12$ labels per run
(Table~\ref{tab:ilc-conv}) and applying no forward correction, so it changes
little either way ($+0.16/+0.74$). This is the appropriate outcome when a local
estimate adds little to a global one. Taken together, \textsc{Cora}
(too fragmented for a reliable local estimate) and \textsc{PubMed} (clean enough
that a local estimate has little to add) delimit the range in which \methodname{}
operates: it improves accuracy where the label-free estimator provides dense,
well-calibrated cluster signal, and neither helps nor harms where it does not.

\paragraph{Robustness of the agreement signal to correlated mislabeling.}
Because homophilous graph neighbours and feature-space neighbours can share the
\emph{same} LLM mistake, a mislabeled node might still attract high agreement and
be scored reliable. We test this directly: on each probe we split seeds by
whether the LLM label matches ground truth (measurement only) and compare mean
agreement (Table~\ref{tab:agree-bias}). The signal stays discriminative on every
benchmark: correctly-labeled seeds attract $14$--$54$pp more neighbour agreement
than mislabeled ones (\textsc{CiteSeer} $79$ vs.\ $25$, \textsc{Cora} $69$ vs.\
$43$, \textsc{PubMed} $87$ vs.\ $67$, \textsc{WikiCS} $48$ vs.\ $21$, \textsc{DBLP}
$55$ vs.\ $41$), so correlated mislabeling does not invert the estimate. Its
residual effect is nonetheless real and largest where noise is \emph{spatially
coherent}. On \textsc{Cora}, whole clusters are mislabeled the same way: a cluster
that is $96\%$ true class-1 is annotated $0\%$ correctly, yet its members agree
with their neighbours almost unanimously, so $T_c$ over-reads reliability there.
This is reflected in \textsc{Cora}'s lowest per-cluster calibration ($r=0.41$,
Table~\ref{tab:casestudy}) and in the \textsc{Cora}-GCN deficit.

\begin{table}[h]
\centering
\small
\setlength{\tabcolsep}{5pt}
\renewcommand{\arraystretch}{1.0}
\begin{tabular}{@{}lccc@{}}
\toprule
\textbf{Dataset} & \textbf{agree\,$|$\,correct} & \textbf{agree\,$|$\,wrong} & \textbf{gap} \\
\midrule
\textsc{CiteSeer} & $78.6$ & $24.7$ & $+53.9$ \\
\textsc{Cora}     & $68.6$ & $42.7$ & $+25.9$ \\
\textsc{PubMed}   & $86.8$ & $66.7$ & $+20.2$ \\
\textsc{WikiCS}   & $47.7$ & $20.6$ & $+27.1$ \\
\textsc{DBLP}     & $54.8$ & $41.1$ & $+13.7$ \\
\bottomrule
\end{tabular}
\caption{\textbf{Agreement signal vs.\ correlated mislabeling.} Mean
cross-modality agreement ($\%$) on probe seeds with correct vs.\ wrong LLM
labels (ground truth used for measurement only; GCN).}
\label{tab:agree-bias}
\end{table}

\section{Estimation on \textsc{ogbn-arxiv}: Budget and Partition Sweep}
\label{app:arxiv}

§\ref{sec:deployment} places \textsc{ogbn-arxiv} outside \methodname{}'s
operating regime on two diagnostics. Since budget and partition size are
under the practitioner's control---unlike the silhouette---we ask
whether either resolves the failure. We re-ran the deployed estimator of
Eq.~\ref{eq:tc-agreement} across annotation budgets and
$K \in \{C, 2C\}$, measuring the low-support fraction, the cluster-level
calibration $r$ (correlation between estimated self-trust and
ground-truth cluster purity), and the cell-level calibration
$r_{\text{cell}}$. Ground-truth labels are used only to score the
estimate.

Both levers help at the cluster level: at $K{=}2C$, $r$ rises from
$0.36$ to $0.48$ as the budget quadruples, and a coarser partition
concentrates the same probes into fewer cells. Neither closes the gap.
At four times the protocol budget, $89$--$94\%$ of cells remain
unsupported and $r_{\text{cell}} \leq 0.21$---and $r_{\text{cell}}$ is
the resolution the pipeline actually consumes. Reducing $K$ also trades
against the effect \methodname{} exploits, since coarser clusters dilute
within-class structure, the same trade visible in the $K$-sweep of
Figure~\ref{fig:ksens}. The failure is therefore one of support rather
than of the phenomenon: cluster-conditional noise on this graph is
present and pronounced, and what is missing is the evidence to localise
it. Whether a better-supported estimate would yield downstream accuracy
gains here is left to future work.

\begin{table}[h]
\centering
\small
\setlength{\tabcolsep}{5pt}
\renewcommand{\arraystretch}{1.1}
\begin{tabular}{@{}llcccc@{}}
\toprule
$K$ & $B$ & probes/cell & low-supp. (\%) & $r$ & $r_{\text{cell}}$ \\
\midrule
$2C{=}80$ & $50C$  & 0.25 & 97 & 0.36 & $\approx 0$ \\
$2C{=}80$ & $100C$ & 0.50 & 96 & 0.45 & $\approx 0$ \\
$2C{=}80$ & $200C$ & 1.00 & 94 & 0.48 & 0.10 \\
$C{=}40$  & $50C$  & 0.50 & 95 & 0.47 & $\approx 0$ \\
$C{=}40$  & $100C$ & 1.00 & 93 & 0.51 & 0.13 \\
$C{=}40$  & $200C$ & 2.00 & 90 & 0.54 & 0.21 \\
\bottomrule
\end{tabular}
\caption{\textbf{Budget and partition sweep on \textsc{ogbn-arxiv}}
(first row is the deployed configuration). Larger budgets and coarser
partitions improve cluster-level calibration $r$, but the
(cluster, class) grid stays under-supported throughout and cell-level
calibration never becomes usable.}
\label{tab:arxiv}
\end{table}

\section{Annotator Robustness: \textsc{gpt-4o-mini}}
\label{app:gpt4omini}

§\ref{sec:exp-annotator} reports the frontier annotator. Table~\ref{tab:gpt4omini}
gives the same experiment under \textsc{gpt-4o-mini}, a
same-generation alternative to \textsc{gpt-3.5-turbo}, with \llmgnn{}
included. \methodname{} leads on seven of the ten cells, with its
largest margins on the GAT cells---the same distribution as the main
results.

The single cell where \methodname{} trails is \textsc{DBLP}-GAT
($75.51$ vs.\ $76.63$, within the $2.05$ seed standard deviation), and
its cause is consistent with our thesis. \textsc{gpt-4o-mini} both
raises and flattens annotation quality on \textsc{DBLP}, compressing
the cluster-conditional structure \methodname{} exploits, while GAT's
attention already discounts inconsistent edges and partially substitutes
for the admission gate. Where the local structure the method models is
absent, it has nothing to add. Under the frontier annotator the same
cell reverses in \methodname{}'s favour ($+0.28$), also within seed
noise, which is what convergence rather than a systematic deficit
predicts.

\begin{table}[h]
\centering
\vspace{-5pt}
\small
\setlength{\tabcolsep}{6pt}
\renewcommand{\arraystretch}{1.1}
\begin{tabular}{@{}lccc@{}}
\toprule
\textbf{Dataset} & \llmgnn{} & \locle{} & \methodname{} \\
\midrule
\rowcolor{grouphdr}\multicolumn{4}{@{}l}{\textit{GAT backbone}}\\
\textsc{CiteSeer} & \val{66.30}{0.38} & \val{68.31}{1.36}  & \valb{74.28}{0.67} \\
\textsc{Cora}     & \val{67.64}{1.99} & \val{72.63}{0.84}  & \valb{75.61}{0.83} \\
\textsc{PubMed}   & \val{74.76}{0.89} & \valb{80.08}{0.66} & \val{79.52}{2.04} \\
\textsc{WikiCS}   & \val{68.45}{0.99} & \val{70.97}{1.76}  & \valb{75.13}{1.03} \\
\textsc{DBLP}     & \val{74.50}{0.49} & \valb{76.63}{0.05} & \val{75.51}{2.05} \\
\midrule
\rowcolor{grouphdr}\multicolumn{4}{@{}l}{\textit{GCN backbone}}\\
\textsc{CiteSeer} & \val{67.37}{0.68} & \val{73.59}{0.05}  & \valb{74.87}{0.82} \\
\textsc{Cora}     & \val{68.97}{0.88} & \valb{76.55}{0.65} & \val{75.04}{0.48} \\
\textsc{PubMed}   & \val{77.03}{0.84} & \val{78.60}{1.18}  & \valb{79.14}{1.85} \\
\textsc{WikiCS}   & \val{70.22}{0.61} & \val{72.87}{0.14}  & \valb{73.52}{0.53} \\
\textsc{DBLP}     & \valb{76.00}{0.29} & \val{75.09}{1.47} & \val{75.90}{1.87} \\
\bottomrule
\end{tabular}
\caption{\textbf{Annotator robustness under \textsc{gpt-4o-mini}}
(test accuracy $\%$, mean${}_{\pm\text{std}}$, 5 seeds). \textbf{Bold} =
best per row. \methodname{} leads on seven of ten cells under a
same-generation alternative to the \textsc{gpt-3.5-turbo} annotator of
Table~\ref{tab:main}; the frontier annotator is in
§\ref{sec:exp-annotator}.}
\label{tab:gpt4omini}
\vspace{-5pt}
\end{table}

\end{document}